\documentclass[10pt,twocolumn,letterpaper]{article}

\usepackage[pagenumbers]{cvpr} 

\usepackage[dvipsnames]{xcolor}
\definecolor{citecolor}{HTML}{0071bc}
\definecolor{color_ao}{gray}{0.5}
\definecolor{color_our}{HTML}{e6f2c2}
\definecolor{color_pre}{rgb}{0.52,0.59,0.69}
\definecolor{Gray}{gray}{0.9}
\definecolor{LighterGray}{gray}{0.93}
\definecolor{LightGrayForTableRule}{gray}{0.92}
\definecolor{DarkGray}{gray}{0.5}
\definecolor{Black}{rgb}{0.0, 0.0, 0.0}
\definecolor{NiceBlue}{rgb}{0.11764705882352941, 0.5647058823529412, 1.0}
\definecolor{NiceGreen}{HTML}{b6c783}
\usepackage{diagbox}
\definecolor{gray}{HTML}{b0aeae}
\definecolor{diffgray}{HTML}{878787}
\definecolor{NiceGray}{HTML}{696969}
\definecolor{applegreen}{rgb}{0.55, 0.71, 0.0}

\definecolor{demphcolor}{RGB}{144,144,144}

\newcount\Comments 
\usepackage{array,multirow,graphicx}
\usepackage{color}
\usepackage{wrapfig}
\definecolor{purple}{rgb}{1,0,1}
\newcolumntype{a}{>{\columncolor{lightblue}}c}
\newcommand{\kibitz}[2]{\ifnum\Comments=1\textcolor{#1}{#2}\fi}

\usepackage{tkz-kiviat}
\usepackage{xfp}
\usetikzlibrary{arrows}

\usepackage{epsfig}
\usepackage{graphicx}
\usepackage{tikz}
\usepackage{pgfplots}
    \pgfplotsset{compat=1.18}
\usepackage{caption}
\usepgfplotslibrary{groupplots}
\usetikzlibrary{patterns}

\usepackage{lipsum}

\definecolor{cvprblue}{rgb}{0.21,0.49,0.74}
\usepackage[pagebackref,breaklinks,colorlinks,citecolor=cvprblue]{hyperref}
\usepackage{multirow}
\usepackage{xcolor,colortbl}
\hypersetup{
    colorlinks,
    linkcolor={red},
    citecolor={green}
}
\usepackage{amssymb}
\usepackage{pifont}
\usepackage{amsmath}
\usepackage{algorithm}
\usepackage{algpseudocode}
\usepackage[rightcaption]{sidecap}
\definecolor{Light}{HTML}{f6fae4}
\definecolor{Light}{HTML}{fafced}
\definecolor{Light}{HTML}{f8fbe9}

\definecolor{6ec2b5}{HTML}{b8d65e}
\definecolor{FFABA8}{HTML}{807d7d}
\colorlet{Dark}{6ec2b5}
\colorlet{Salmon}{FFABA8}

\newcommand{\CC}[1]{\cellcolor{Light}}

\newcommand{\model}{\text{VistaLLM}}
\newcommand{\data}{\text{CoinIt}}

\newcommand\blfootnote[1]{%
  \begingroup
  \renewcommand\thefootnote{}\footnote{#1}%
  \addtocounter{footnote}{-1}%
  \endgroup
}

\def\paperID{10850} 
\def\confName{CVPR}
\def\confYear{2024}

\title{Jack of All Tasks, Master of Many: Designing General-purpose \\ Coarse-to-Fine Vision-Language Model}

\author{Shraman Pramanick$^{*1,2 \dagger}$ \ \ \ \  Guangxing Han$^{*2}$  \ \ \ \ Rui Hou$^{2}$ \ \ \ \  Sayan Nag$^{3}$ \ \ \ \  Ser-Nam Lim$^{4}$ \\ Nicolas Ballas$^{2}$ \ \ \ \ Qifan Wang$^{2}$  \ \ \ \ Rama Chellappa$^{1}$ \ \ \ \ Amjad Almahairi$^{2}$\vspace{1em}\\
$^{1}$Johns Hopkins University, $^{2}$Meta, $^{3}$University of Toronto, $^{4}$University of Central Florida \\
}

\begin{document}
\maketitle

\begin{abstract}

\vspace{-2mm}
The ability of large language models (LLMs) to process visual inputs has given rise to general-purpose vision systems, unifying various vision-language (VL) tasks by instruction tuning. However, due to the enormous diversity in input-output formats in the vision domain, existing general-purpose models fail to successfully integrate segmentation and multi-image inputs with coarse-level tasks into a single framework. In this work, we introduce \model, a powerful visual system that addresses coarse- and fine-grained VL tasks over single and multiple input images using a unified framework. \model\ utilizes an instruction-guided image tokenizer that filters global embeddings using task descriptions to extract compressed and refined features from numerous images. Moreover, \model\ employs a gradient-aware adaptive sampling technique to represent binary segmentation masks as sequences, significantly improving over previously used uniform sampling. To bolster the desired capability of \model, we curate \data, a comprehensive coarse-to-fine instruction tuning dataset with $6.8$M samples. We also address the lack of multi-image grounding datasets by introducing a novel task, AttCoSeg (Attribute-level Co-Segmentation), which boosts the model's reasoning and grounding capability over multiple input images. Extensive experiments on a wide range of V- and VL tasks demonstrate the effectiveness of \model\ by achieving consistent state-of-the-art performance over strong baselines across many downstream tasks. Our project page can be found at \href{https://shramanpramanick.github.io/VistaLLM/}{https://shramanpramanick.github.io/VistaLLM/}.
\blfootnote{$^*$Equal technical contribution.}
\blfootnote{$^\dagger$Part of this work was done during an internship at Meta.}
\end{abstract}

\vspace{-10mm}
\section{Introduction}

\begin{figure}[!t]
\vspace{8pt}
\resizebox{1.095\linewidth}{!}{
    \hspace{-88pt}
    \newcommand{\lattice}{4}
\newcommand{\naxis}{14}

\newcommand{\amax}{53.0}
\newcommand{\agpt}{51.9}

\newcommand{\bmax}{47.9}
\newcommand{\bgpt}{47.7}

\newcommand{\cmax}{60.1}
\newcommand{\cgpt}{58.2}

\newcommand{\dmax}{89.9}
\newcommand{\dshikra}{87.8}
\newcommand{\dgpt}{88.7}
\newcommand{\dferret}{89.5}

\newcommand{\emax}{90.5}
\newcommand{\eshikra}{86.2}
\newcommand{\eferret}{89.8}

\newcommand{\fmax}{76.9}
\newcommand{\fshikra}{75.3}

\newcommand{\gmax}{128.4}
\newcommand{\gshikra}{117.5}

\newcommand{\hmax}{85.5}
\newcommand{\hshikra}{85.3}
\newcommand{\hgptroi}{84.8}

\newcommand{\imax}{79.1}
\newcommand{\igptroi}{78.6}

\newcommand{\jmax}{54.6}

\newcommand{\kmax}{77.2}

\newcommand{\lmax}{65.1}

\newcommand{\mmax}{95.1}

\newcommand{\nmax}{81.3}

\newcommand{\origin}{0.92} 
\newcommand{\horigin}{0.985}
\newcommand{\dorigin}{0.965}
\newcommand{\gorigin}{0.78}
\newcommand{\iorigin}{0.96}
\newcommand{\forigin}{0.945}

\newcommand\ColorBox[1]{\textcolor{#1}{\rule{3ex}{3ex}}}

\newcommand{\annotMark}[5]{
	\pgfmathsetmacro{\xcor}{#3*cos{(#1*#2)}/(1/#4)};
	\pgfmathsetmacro{\ycor}{#3*sin{(#1*#2)}/(1/#4)};
	\draw (\xcor,\ycor)node[anchor=south]{\large #5};
}

\begin{tikzpicture}[rotate=0, scale=1.0,every node/.style={inner sep=-15,outer sep=-15}]
	\tkzKiviatDiagram[lattice=\lattice, gap=1.0, step=1, label space=1.5]
	{   \textbf{\textcolor{Red}{TextVQA}},
        \textbf{\textcolor{Plum}{IconQA}},
        \textbf{\textcolor{Red}{HM}},
        \textbf{\textcolor{blue}{REC}},
		\textbf{\textcolor{Red}{POPE}},
		\textbf{\textcolor{Red}{VQAv2}},
		\textbf{\textcolor{Red}{COCO Cap}},
        \textbf{\textcolor{blue}{BoxQA}},
		\textbf{\textcolor{Red}{VCR}},
		\textbf{\textcolor{blue}{GREC}},
		\textbf{\textcolor{blue}{RES}},
		\textbf{\textcolor{blue}{GRES}},
        \textbf{\textcolor{Plum}{iCoSeg}},
        \textbf{\textcolor{OliveGreen}{NLVR2}}
        }
		\tkzKiviatLine[thick, fill=color_our!80, color=NiceGreen, opacity=0.7](
		\fpeval{(\amax/\amax-\origin)/(1 - \origin)*\lattice},
		\fpeval{(\bmax/\bmax-\origin)/(1 - \origin)*\lattice},
		\fpeval{(\cmax/\cmax-\origin)/(1 - \origin)*\lattice},
		\fpeval{(\dmax/\dmax-\dorigin)/(1 - \dorigin)*\lattice},
		\fpeval{(\emax/\emax-\origin)/(1 - \origin)*\lattice},
		\fpeval{(\fmax/\fmax-\forigin)/(1 - \forigin)*\lattice},
		\fpeval{(\gmax/\gmax-\gorigin)/(1 - \gorigin)*\lattice},
		\fpeval{(\hmax/\hmax-\horigin)/(1 - \horigin)*\lattice},
        \fpeval{(\imax/\imax-\iorigin)/(1 - \iorigin)*\lattice},
		\fpeval{(\jmax/\jmax-\origin)/(1 - \origin)*\lattice},
        \fpeval{(\kmax/\kmax-\origin)/(1 - \origin)*\lattice},
		\fpeval{(\lmax/\lmax-\origin)/(1 - \origin)*\lattice},
        \fpeval{(\mmax/\mmax-\origin)/(1 - \origin)*\lattice},
		\fpeval{(\nmax/\nmax-\origin)/(1 - \origin)*\lattice}
        )
        \tkzKiviatLine[thick, fill=Periwinkle!100, color=Periwinkle, opacity=0.6](
		\fpeval{(\origin-\origin)/(1 - \origin)*\lattice},
		\fpeval{(\origin-\origin)/(1 - \origin)*\lattice},
		\fpeval{(\origin-\origin)/(1 - \origin)*\lattice},
		\fpeval{(\dferret/\dmax-\dorigin)/(1 - \dorigin)*\lattice},
		\fpeval{(\eferret/\emax-\origin)/(1 - \origin)*\lattice},
        \fpeval{(\forigin-\forigin)/(1 - \forigin)*\lattice},
		\fpeval{(\gorigin-\gorigin)/(1 - \gorigin)*\lattice},
		\fpeval{(\horigin-\horigin)/(1 - \horigin)*\lattice},
        \fpeval{(\iorigin-\iorigin)/(1 - \origin)*\lattice},
		\fpeval{(\origin-\origin)/(1 - \origin)*\lattice},
        \fpeval{(\origin-\origin)/(1 - \origin)*\lattice},
		\fpeval{(\origin-\origin)/(1 - \origin)*\lattice},
        \fpeval{(\origin-\origin)/(1 - \origin)*\lattice},
		\fpeval{(\origin-\origin)/(1 - \origin)*\lattice}
        )
        \tkzKiviatLine[thick, fill=BurntOrange!80, color=BurntOrange, opacity=0.7](
		\fpeval{(\origin-\origin)/(1 - \origin)*\lattice},
		\fpeval{(\origin-\origin)/(1 - \origin)*\lattice},
		\fpeval{(\origin-\origin)/(1 - \origin)*\lattice},
		\fpeval{(\dshikra/\dmax-\dorigin)/(1 - \dorigin)*\lattice},
		\fpeval{(\eshikra/\emax-\origin)/(1 - \origin)*\lattice},
        \fpeval{(\fshikra/\fmax-\forigin)/(1 - \forigin)*\lattice},
		\fpeval{(\gshikra/\gmax-\gorigin)/(1 - \gorigin)*\lattice},
		\fpeval{(\hshikra/\hmax-\horigin)/(1 - \horigin)*\lattice},
        \fpeval{(\iorigin-\iorigin)/(1 - \origin)*\lattice},
		\fpeval{(\origin-\origin)/(1 - \origin)*\lattice},
        \fpeval{(\origin-\origin)/(1 - \origin)*\lattice},
		\fpeval{(\origin-\origin)/(1 - \origin)*\lattice},
        \fpeval{(\origin-\origin)/(1 - \origin)*\lattice},
		\fpeval{(\origin-\origin)/(1 - \origin)*\lattice}
        )
        \tkzKiviatLine[thick, fill=NiceGray!100, color=NiceGray!100, opacity=0.9](
		\fpeval{(\origin-\origin)/(1 - \origin)*\lattice},
		\fpeval{(\origin-\origin)/(1 - \origin)*\lattice},
		\fpeval{(\origin-\origin)/(1 - \origin)*\lattice},
		\fpeval{(\dorigin-\dorigin)/(1 - \dorigin)*\lattice},
		\fpeval{(\origin-\origin)/(1 - \origin)*\lattice},
        \fpeval{(\forigin-\forigin)/(1 - \forigin)*\lattice},
		\fpeval{(\gorigin-\gorigin)/(1 - \gorigin)*\lattice},
		\fpeval{(\hgptroi/\hmax-\horigin)/(1 - \horigin)*\lattice},
        \fpeval{(\igptroi/\imax-\iorigin)/(1 - \iorigin)*\lattice},
		\fpeval{(\origin-\origin)/(1 - \origin)*\lattice},
        \fpeval{(\origin-\origin)/(1 - \origin)*\lattice},
		\fpeval{(\origin-\origin)/(1 - \origin)*\lattice},
        \fpeval{(\origin-\origin)/(1 - \origin)*\lattice},
		\fpeval{(\origin-\origin)/(1 - \origin)*\lattice}
        )
        \tkzKiviatLine[thick, fill=Maroon!70, color=Maroon!60, opacity=0.8](
		\fpeval{(\agpt/\amax-\origin)/(1 - \origin)*\lattice},
		\fpeval{(\bgpt/\bmax-\origin)/(1 - \origin)*\lattice},
		\fpeval{(\cgpt/\cmax-\origin)/(1 - \origin)*\lattice},
		\fpeval{(\dgpt/\dmax-\dorigin)/(1 - \dorigin)*\lattice},
		\fpeval{(\origin-\origin)/(1 - \origin)*\lattice},
        \fpeval{(\forigin-\forigin)/(1 - \forigin)*\lattice},
		\fpeval{(\gorigin-\gorigin)/(1 - \gorigin)*\lattice},
		\fpeval{(\horigin-\horigin)/(1 - \horigin)*\lattice},
        \fpeval{(\iorigin-\iorigin)/(1 - \iorigin)*\lattice},
		\fpeval{(\origin-\origin)/(1 - \origin)*\lattice},
        \fpeval{(\origin-\origin)/(1 - \origin)*\lattice},
		\fpeval{(\origin-\origin)/(1 - \origin)*\lattice},
        \fpeval{(\origin-\origin)/(1 - \origin)*\lattice},
		\fpeval{(\origin-\origin)/(1 - \origin)*\lattice}
        )
        
	\annotMark{0.15}{360/\naxis}{4.35}{1}{\amax};
	\annotMark{1.0}{370/\naxis}{4.7}{1}{\bmax};
	\annotMark{1.975}{360/\naxis}{4.8}{1}{\cmax};
	\annotMark{3}{342/\naxis}{4.8}{1}{\dmax};
	\annotMark{4.18}{360/\naxis}{4.8}{1}{\emax};
	\annotMark{5.05}{360/\naxis}{4.8}{1}{\fmax};
	\annotMark{6}{360/\naxis}{4.8}{1}{\gmax};
	\annotMark{7}{353/\naxis}{4.36}{1}{\hmax};
	\annotMark{8}{358/\naxis}{4.15}{1}{\imax};
	\annotMark{9.1}{358/\naxis}{3.9}{1}{\jmax};
    \annotMark{10.25}{358/\naxis}{3.7}{1}{\kmax};
	\annotMark{11.4}{358/\naxis}{3.8}{1}{\lmax};
    \annotMark{12.5}{358/\naxis}{4.1}{1}{\mmax};
	\annotMark{13.5}{358/\naxis}{4.3}{1}{\nmax};

    \annotMark{3}{325/\naxis}{2.3}{1}{\dshikra};
	\annotMark{4}{383/\naxis}{2.7}{1}{\eshikra};
	\annotMark{5}{365/\naxis}{3.3}{1}{\fshikra};
	\annotMark{6}{360/\naxis}{3.4}{1}{\gshikra};
	\annotMark{7}{350/\naxis}{3.2}{1}{\hshikra};
    \annotMark{7}{315/\naxis}{1.95}{1}{\hgptroi};
	\annotMark{8}{357/\naxis}{3.0}{1}{\igptroi};
    \annotMark{0.2}{360/\naxis}{3.2}{1}{\agpt};
	\annotMark{1.0}{400/\naxis}{3.29}{1}{\bgpt};
	\annotMark{1.975}{360/\naxis}{3.3}{1}{\cgpt};
	\annotMark{3}{335/\naxis}{3.3}{1}{\dgpt};
    \annotMark{3}{340/\naxis}{4.0}{1}{\dferret};
	\annotMark{4}{375/\naxis}{4.0}{1}{\eferret};
	\node[anchor=south west,xshift=25pt,yshift=26pt] at (current bounding box.south east)
{
	\begin{tabular}{@{}lp{5cm}@{}}
        \ColorBox{color_our!100} & \large \model-13B  \\
        \ColorBox{Maroon!100} & \large MiniGPT-v2 \\
        \ColorBox{Periwinkle!100} & \large Ferret-13B \\
        \ColorBox{BurntOrange!100} & \large Shikra-13B \\
        \ColorBox{NiceGray!100} & \large GPT4RoI-13B \\
	\end{tabular}
 };
    \node[anchor=north east,xshift=39pt,yshift=-24.5pt] at (current bounding box.north east)
{
	\begin{tabular}{@{}lp{7cm}@{}}
        \tikz\draw[Red!100,fill=Red!100] (0,0) circle (1.5ex); & \large Single-image coarse-level \\
        \tikz\draw[blue!100,fill=blue!100] (0,0) circle (1.5ex); & \large Single-image region-level\\
		\tikz\draw[OliveGreen!100,fill=OliveGreen!100] (0,0) circle (1.5ex); & \large Multi-image coarse-level\\
        
        \tikz\draw[Plum!100,fill=Plum!100] (0,0) circle (1.5ex); & \large Multi-image region-level\\
	\end{tabular}
};
\end{tikzpicture}%
}

\vspace{-1mm}
\caption{\textbf{\model\ achieves the state-of-the-art} performance across a broad range of single and multi-image coarse-to-fine grained reasoning and grounding tasks (see Table \ref{tab:dataset_details} for details) among general-purpose baselines. Notably, no existing baseline have unified segmentation and multi-image tasks in a single system. We show officially reported numbers for every baseline.}
\label{fig:results_summary}
\vspace{-6mm}
\end{figure}
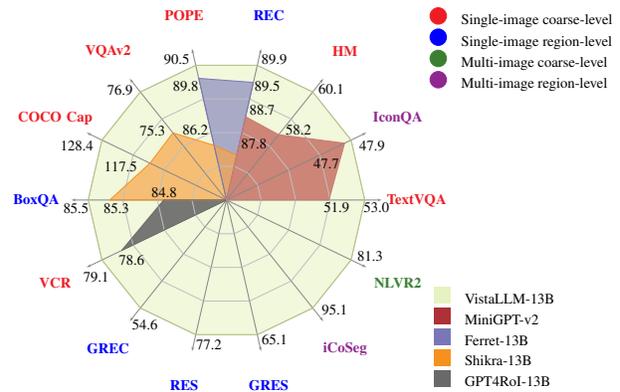

\vspace{-1mm}
Large language models (LLM) have proven to be the \textit{de-facto} solution to address novel natural language processing (NLP) tasks, thanks to their ability to comprehend user-tailored prompts, instructions, and detailed task descriptions \cite{openai2022chatgpt, chowdhery2022palm, hoffmann2022empirical, touvron2023llama, touvron2023llama2, penedo2023refinedweb}. However, the problem is more challenging the vision domain due to an inherent disparity of input and output formats across different tasks. Though pre-training followed by a fine-tuning strategy is effective for various vision problems \cite{radford2021learning, dou2022empirical, dou2022coarse, yang2022unitab, li2022grounded, cheng2023vindlu, pramanick2023volta, li2023blip2, beit2023, flip2023, pramanick2023egovlpv2, lin2023univtg, cheng2024dam}, with the continuously increasing model parameters, the marginal cost for task-specific tuning comes with significant computational overhead. Hence, it becomes crucial to design general-purpose vision models that can perceive natural-language instructions to solve various vision problems in a zero-shot manner.

The development of general-purpose vision models faces two significant challenges: first, the unification of diverse input-output formats, and second, an effective representation of visual features for a variety of tasks. Image-level vision tasks such as classification, captioning, and question-answering involve textual outputs and primarily require a broader, coarse-grained image representation, making them relatively straightforward to integrate into a unified framework \cite{liu2023llava, zhu2023minigpt, instructblip, gao2023llama}. In contrast, region-level prediction tasks like object detection and semantic segmentation necessitate fine-grained, pixel-scale visual features and produce dense outputs such as bounding boxes and masks. Converting bounding boxes to natural language sequences is feasible by serializing the coordinates of two corners. However, representing a binary mask as a text sequence poses a more complex challenge, especially when dealing with multiple input images each associated with numerous segmentation masks. Although some recent general-purpose systems have succeeded in unifying coarse-level tasks with object detection \cite{chen2023shikra, zhang2023gpt4roi, you2023ferret, chen2023minigptv2, peng2023kosmos2, jiang2023comm}, they do not incorporate segmentation within the same framework. Furthermore, the capabilities of these existing systems are often limited to processing single-image input, thereby constraining their applicability in broader, more complex scenarios, such as reasoning over multiple images and recognizing and segmenting common objects.

In this work, we present \textbf{\model}, the first general-purpose vision model that addresses coarse- and fine-grained vision-language reasoning and grounding tasks over single and multiple input images. We unify these tasks by converting them into an instruction-following sequence-to-sequence format. We efficiently transform binary masks into a sequence of points by proposing a gradient-aware adaptive contour sampling scheme, which significantly improves over the naive uniform sampling technique previously used for sequence-to-sequence segmentation tasks \cite{chen2021pix2seq, chen2022unified, zhu2022seqtr, liu2023polyformer}. Moreover, to preserve global and region-level information from multiple input images, we propose utilizing a QFormer \cite{li2023blip2} based instruction-guided image tokenizer. Leveraging LLMs' language reasoning ability, we feed our visual features with carefully designed task-specific instructions to LLMs, which generate responses following the instructions. Integrating various tasks with different granularity into such a unified, cohesive, and end-to-end system helps improve the performance of each task by sharing coarse- and fine-grained feature representation.



To train \model\ on a versatile form of vision and language tasks, we collect \textbf{\data} (\textbf{Co}arse-to-f\textbf{in}e \textbf{I}nstruction-\textbf{t}uning Dataset) with $6.8$M samples, ranging over four broad categories of tasks - single-image coarse-level, single-image region-level, multi-image coarse-level, and multi-image region-level. We address the lack of publicly-available multi-image region-level datasets by proposing a novel task, AttCoSeg (\textbf{Att}ribute-level \textbf{Co}-\textbf{Seg}mentation), which aims to recognize input images which have objects with common attributes (shape, color, size, position), and segment those objects. AttCoSeg contains 804k training samples, and help \model\ to gain significant generalizable reasoning and grounding capability over multiple input images. Other tasks of \data\ are constructed by converting publicly available benchmarks into instruction-following format, such as COCO \cite{lin2014microsoft}, Flickr \cite{plummer2015flickr30k}, VCR \cite{zellers2019recognition}, LLaVA \cite{liu2023llava}, VG \cite{krishna2017visual}, PASCAL \cite{faktor2013co} etc. Extensive evaluation on $15$ different benchmarks proves the efficacy of \model, which even surpasses specialist (or fine-tuned) systems in most tasks, including $10.9\%$ CIDEr points gain over Shikra \cite{chen2023shikra} on image captioning, $13.1\%$, $6.7\%$ precision and gIoU improvements over MDETR \cite{kamath2021mdetr} on GREC and GRES, $3\%$ $\mathcal{J}$-index gains over CycleSegNet \cite{zhang2021cyclesegnet} on iCoSeg. 


In summary, our contributions are threefold: $(i)$We propose \model, equipped with a instruction-guided image tokenizer, to seamlessly integrate coarse- and fine-grained vision-language reasoning and grounding tasks over single and multiple input images into a unified general-purpose model. $(ii)$ To efficiently convert segmentation masks into a sequence, we propose a gradient-aware adaptive contour sampling scheme, which improves over previously used uniform sampling by $3-4$ mIoU scores on different segmentation benchmarks. $(iii)$ We construct \data, a large-scale coarse-to-fine instruction-tuning dataset, for model training. Moreover, we introduce a novel task, AttCoSeg, which addresses the lack of publicly available multi-image grounding datasets. We evaluate \model\ on a wide-range of vision-language tasks across $15$ benchmarks, achieving state-of-the-art performance in all of them, even surpassing specialist systems. We summarize these results in Figure \ref{fig:results_summary}.


\vspace{-1mm}
\section{Related Works}
\vspace{-1mm}

General-purpose vision models, also known as multimodal large language models (MLLM), have recently been proven to be an effective way to unify a versatile array of vision and language tasks. These models, which use potent LLMs \cite{openai2022chatgpt, chowdhery2022palm, hoffmann2022empirical, touvron2023llama, touvron2023llama2, penedo2023refinedweb, workshop2022bloom, zhang2022opt, taylor2022galactica, iyer2022optiml, zeng2022glm, borgeaud2022improving, du2022glam, victor2022multitask, wei2021finetuned} to reason textual instructions, can broadly be categorized into two groups based on their input and output formats:

\vspace{1mm}
\noindent \textbf{Coarse-level MLLMs:} Early attempts of designing MLLMs focused on image-level vision tasks with textual outputs, such as visual question answering \cite{antol2015vqa, singh2019towards, lu2021iconqa, hudson2019gqa} and image captioning \cite{herdade2019image, hossain2019comprehensive}. Frozen \cite{tsimpoukelli2021multimodal}, Flamingo \cite{alayrac2022flamingo}, FrozenBiLM \cite{yang2022zero}, MAGMA \cite{eichenberg2022magma}, ClipCap \cite{mokady2021clipcap}, VidIL \cite{wang2022language}, PICa \cite{yang2022empirical} are among the first few to show the in-context capability of LLMs for few-shot vision tasks. More recent works have focused on using LLMs for visual instruction tuning. To name a few, LLaVA \cite{liu2023llava}, MiniGPT-4 \cite{zhu2023minigpt}, MM-REACT \cite{yang2023mm}, BLIP2 \cite{li2023blip2}, mPLUS-OWL \cite{ye2023mplug}, LLaMA-Adapter v2 \cite{gao2023llama}, Otter \cite{li2023otter}, Instruct-BLIP \cite{instructblip}, LLaVA-Med  \cite{li2023llavamed} have been proven to be effective. However, these models lack region-specific capabilities and can not perform visual grounding tasks.  

\begin{figure*}
\centering
\includegraphics[width=0.94\textwidth]{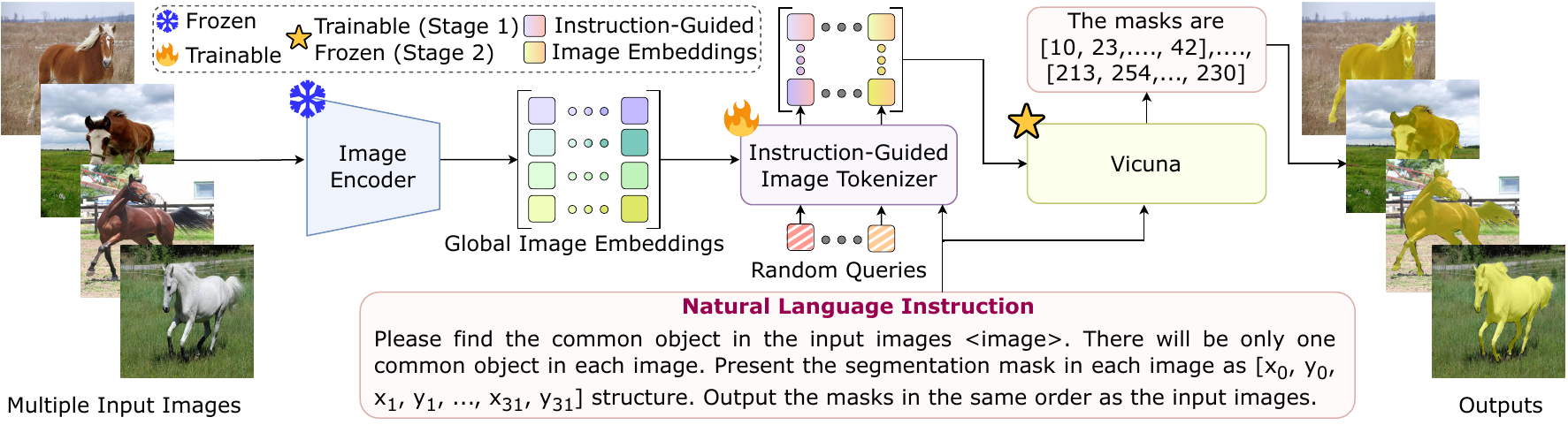}
\vspace{-2mm}
\caption{\textbf{Overview of the proposed system - \model}, which integrates single- and multi-image coarse- and fine-grained vision-language tasks into a unified general-purpose framework. \model\ contains three key design modules - ($i$) image encoder to extract the global image embedding, ($ii$) instruction-guided image tokenizer, which refines and compresses the global image embeddings using task instruction, enabling the model to filter the necessary visual information required for the current task, and ($iii$) LLM (Vicuna)-based decoder to jointly process image and language features, and generate the desired output. \model\ uses a gradient-aware adaptive sampling technique to efficiently represent segmentation masks as a point sequence, described in Section \ref{sec:sequence_generation}. All parameters except the image encoder are trained in stage 1, while only the image tokenizer is fine-tuned in stage 2 (See Section \ref{sec:model_architecture}, \ref{sec:implementation_details} for details).}
\label{fig:system_main}
\vspace{-5 mm}
\end{figure*}

\vspace{1mm}
\noindent \textbf{Region-level MLLMs:}
More recently, MLLMs have moved forward to unify region-based referring and grounding tasks into general-purpose vision systems. KOSMOS-2 \cite{peng2023kosmos2}, VisionLLM \cite{wang2023visionllm}, Shikra \cite{chen2023shikra}, GPT4RoI \cite{zhang2023gpt4roi}, All-Seeing Model \cite{wang2023all}, CogVLM \cite{wang2023cogvlm}, COMM \cite{jiang2023clip}, MiniGPT-v2 \cite{chen2023minigptv2} and Ferret \cite{you2023ferret} has shown the capability of MLLMs of fine-grained image comprehension and region-focused conversation. While KOSMOS-2, Shikra, and VisionLLM feed the image coordinates directly into the LLM, GPT4RoI and Ferret use additional feature extractor modules to represent image regions. On a related regime, InternGPT \cite{liu2023internchat}, BuboGPT \cite{zhao2023bubogpt}, and LISA \cite{lai2023lisa} utilize external vision modules to perform grounding tasks. However, these works are only capable of processing single-input images. In this work, we propose \model\ to address all possible reasoning and grounding tasks over single and multiple images. Moreover, we efficiently convert binary masks into sequence by a novel adaptive sampling, which helps to unify segmentation into a general-purpose framework.   

\vspace{-2mm}
\section{Method}
\vspace{-2mm}

We start by presenting the model architecture of \model. Next, we detail the proposed sequence generation approach for segmentation masks and illustrate its efficacy compared to uniform sampling.
\vspace{-1mm}
\subsection{Model Architecture} \label{sec:model_architecture}
\vspace{-1mm}

The overall architecture of \model, shown in Figure \ref{fig:system_main}, consists of three key design modules - ($i$) image encoder to extract the global image embedding, ($ii$) instruction-guided image tokenizer, which refines and compresses the global image embeddings using task instruction, enabling the model to filter the necessary visual information required for the current task, and ($iii$) LLM-based decoder to jointly process image and language features, and generate the desired output.

\noindent \textbf{Image Encoder.} Given a set of $k$ input images $X=\{x_{\mathrm{i}}\}_{1}^{k}$; $x_{\mathrm{i}} \in \mathbb{R}^{H_i \times W_i \times 3}$, where $H_i$ and $W_i$ denote the height and width of the $i^{\text{th}}$ image, we first feed them into a pre-trained image encoder, EVA-CLIP \cite{sun2023eva}, to extract $k$ image embeddings $Z=\{z_{\mathrm{i}}\}_{1}^{k}$; $z_{\mathrm{i}}\in\mathbb{R}^{N_{\mathrm{i}} \times D}$, $N_i$ is number of spatial tokens in the $i^{\text{th}}$ image and $D$ is the hidden dimension. Note that, for larger $k$, the image feature dimension increases, making it difficult for the LLM decoder to process it as input, which is taken care of in the tokenizer module.   

\begin{figure*}[!t]
\centering
 \begin{subfigure}[b]{0.47\textwidth}
\centering
\includegraphics[width=\textwidth]{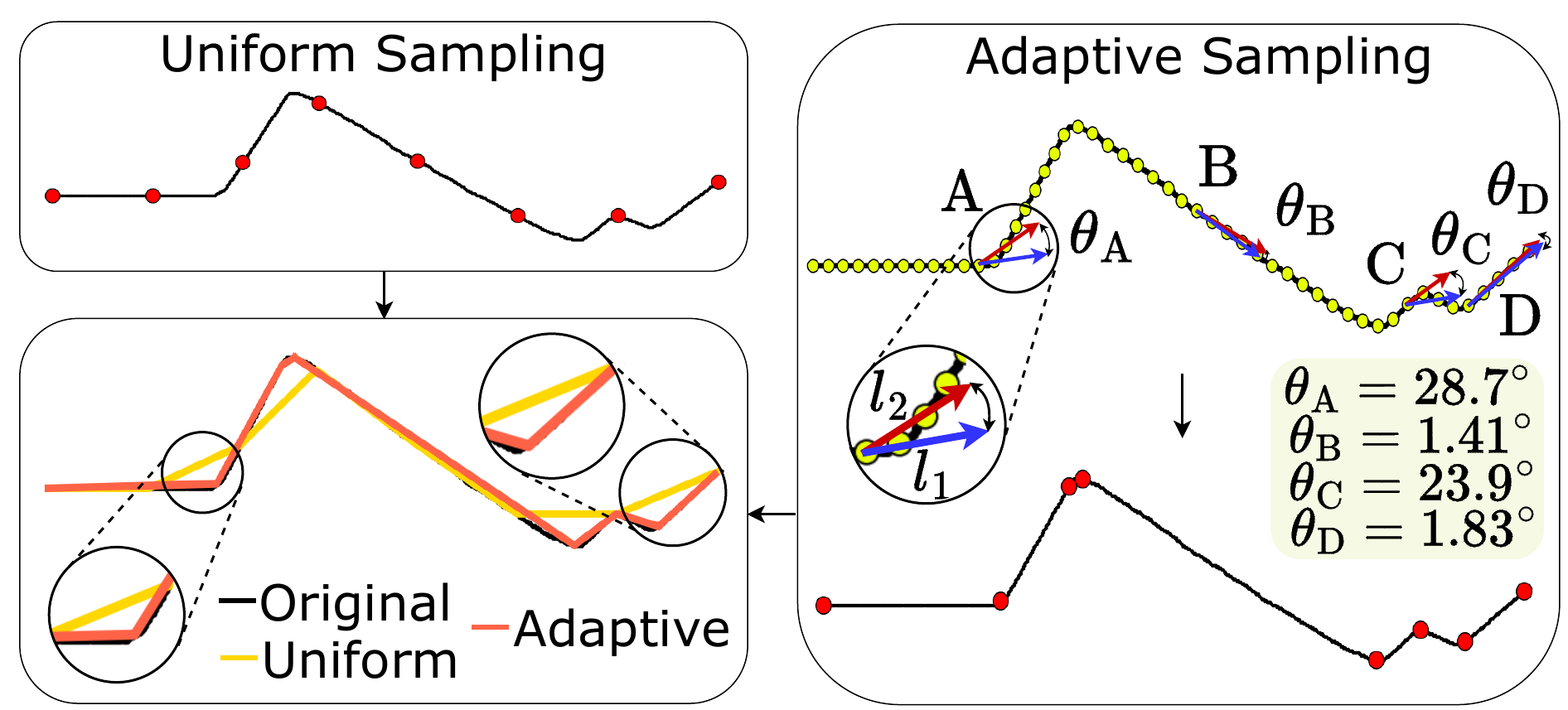}
\caption{\textbf{Illustration of uniform and adaptive sampling} on a line curve.}
\label{fig:sampling_illustration}
 \end{subfigure}
 \begin{subfigure}[b]{0.51\textwidth} 
\centering
\includegraphics[width=\textwidth]{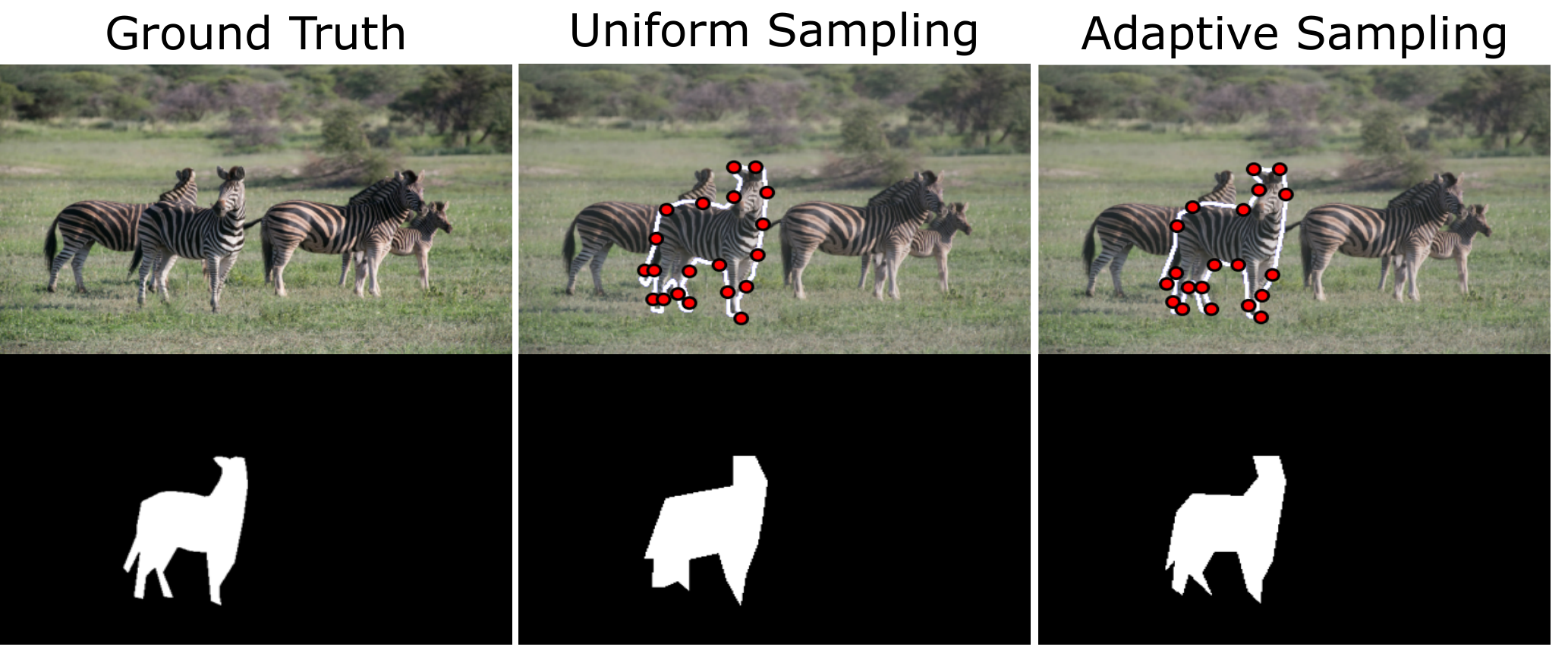}
\caption{\textbf{Illustration of uniform and adaptive sampling} on a object mask.}
\label{fig:sampling_example}
 \end{subfigure}
\vspace{-3mm}
\caption{\textbf{Visualization of uniform and adaptive sampling strategies.} (a) illustration of sampled points and comparison of reassembled curves, (b) illustration of sampled points and comparison of reassembled masks.}
\vspace{-5.5mm}
\label{fig:enter-label}
\end{figure*}

\noindent \textbf{Instruction-guided Image Tokenizer.} Unlike many previous general-purpose vision systems \cite{liu2023llava, chen2023shikra, peng2023kosmos2, chen2023minigptv2}, which directly feed the global image features into the decoder, we introduce an instruction-guided image tokenizer, which plays three crucial roles: ($i$) refines the image embeddings in alignment with task description, i.e. for coarse-level tasks, global features are important, whereas for fine-level tasks, only the region features need to be processed. ($ii$) compresses the image embeddings, which is important when there are many input images, and ($iii$) flexibly projects multiple input images with different heights and widths into the same feature dimension. 

The image tokenizer module takes image embeddings and the language instruction and outputs the refined and compressed visual features. If referring regions (points, boxes, masks) are present in the instruction, they are converted to text-interleaved sequence as described in Section \ref{sec:sequence_generation}. Afterwards, we propose to adopt a QFormer \cite{li2023blip2} network with $L$ ($L < N_{\mathrm{i}}, \forall i$) randomly-initialized queries, which learns high-level task-specific information using the language instruction. The output from the tokenizer, $F=\{f_{\mathrm{i}}\}_{1}^{k}$; $f_{\mathrm{i}}\in\mathbb{R}^{L \times D}$, are then flattened to produce the final visual features, $F_{\mathrm{v}} \in\mathbb{R}^{kL \times D}$ which are fed into the LLM. 

\noindent \textbf{LLM.} We use Vicuna \cite{chiang2023vicuna} as our language model, which is a decoder-only LLM \cite{brown2020language} with a context length of $2048$ build by instruction-tuning LLaMa \cite{touvron2023llama}. The LLM takes the vision features $F_{\mathrm{v}}$ and the language instruction as input, and generates task-specific output. We train the LLM end-to-end by traditional next-token prediction objective calculated over the ground-truth. Since Vicuna only has the digits $0$-$9$ in its vocabulary, we introduce additional tokens $10$-$999$ to represent quantized coordinates. During evaluation, we dequantize the generated number tokens into the image space for metric calculation.

\vspace{-1mm}
\subsection{Sequence Generation for Grounding Tasks} \label{sec:sequence_generation}
\vspace{-1mm}

The outputs from grounding tasks typically manifest in one of three formats: points, boxes, and masks. Points and boxes are straightforward to quantify and serialize, as evidenced in \cite{peng2023kosmos2, chen2023shikra, chen2023minigptv2}. For instance, a point is represented by its coordinates [$x$, $y$], while a box is denoted by its diagonal corner points [$x_{\mathrm{min}}$, $y_{\mathrm{min}}$, $x_{\mathrm{max}}$, $y_{\mathrm{max}}$], signifying the top-left and bottom-right corners. Conversely, the outline of a mask can assume any free-form shape comprising potentially infinite points. In scenarios where such free-form polygons are referenced in the input instructions, they can be encoded as region features \cite{zhang2023gpt4roi, you2023ferret}. However, translating segmentation masks into a sequence for output by a general-purpose framework is particularly challenging, and the process necessitates conversion of segmentation masks into a small number of discrete points.

Previously, encoder-decoder-based segmentation approaches \cite{chen2021pix2seq, chen2022unified, zhu2022seqtr, liu2023polyformer} uniformly sample $N$ points clockwise from the contour of the mask, and then quantize and serialize them as [$x_{\mathrm{1}}$, $y_{\mathrm{1}}$, $x_{\mathrm{2}}$, $y_{\mathrm{2}}$, \dots, $x_{\mathrm{N}}$, $y_{\mathrm{N}}$],

\vspace{-5.5mm}
\begin{align}\label{eq:quantize}
x_{\mathrm{i}} = \text{round}\left(\frac{\Tilde{x}_{\mathrm{i}}}{w} * n_{\mathrm{bins}}\right), \;\; y_{\mathrm{i}} = \text{round}\left(\frac{\Tilde{y}_{\mathrm{i}}}{h} * n_{\mathrm{bins}}\right)
\end{align}
\vspace{-4.5mm}

where ($\Tilde{x}_{\mathrm{i}}$,$\Tilde{y}_{\mathrm{i}}$) are the original floating point image coordinates, $w$, $h$ are the width and height of the image, $n_{\mathrm{bins}}$ is the number of quantization bins, and ($x_{\mathrm{i}}$,$y_{\mathrm{i}}$) are the quantized coordinates. However, as shown in the top-left of Figure \ref{fig:sampling_illustration}, the uniform sampling approach is unaware of the contour curvature and cannot properly represent sharp edges. To alleviate this limitation, we argue that the sampling should preserve more points where the contour has a sharp bend and less where it is almost straight. Based on this observation, we propose a gradient-aware adaptive sampling technique, which we describe in three steps:

\vspace{0.5mm}
\begin{itemize}[leftmargin=*]
\item \textbf{Contour Discretization.} First, we discretize the continuous contour by uniformly sampling a high number ($M$) of dense points. Note that these dense points represent the curve well, but such a long sequence is infeasible for training a decoder.
\item \textbf{Gradient Calculation.} Next, for every point $p_{\mathrm{i}\in\{1,\dots,M\}}$ on the curve, we draw two lines - $l_{\mathrm{1}}$ by joining $p_{\mathrm{i}}$ with its previous point $p_{\mathrm{i-1}}$, and $l_{\mathrm{2}}$ by joining $p_{\mathrm{i-1}}$ with the next point $p_{\mathrm{i+1}}$. $l_{\mathrm{1}}$ and $l_{\mathrm{2}}$ create an angle $\theta_{\mathrm{i}}$ ($0^{\circ} \leq \theta_{\mathrm{i}} < 180^{\circ}$) at $p_{\mathrm{i-1}}$. If $\theta_{\mathrm{i}} \simeq 0$, the contour is almost linear at $p_{\mathrm{i}}$, and we can safely discard $p_{\mathrm{i}}$ (e.g., points B and D in the right column of Figure \ref{fig:sampling_illustration}). As $\theta_{\mathrm{i}}$ increases, the curvature at $p_{\mathrm{i}}$ becomes sharper, and the importance of keeping $p_{\mathrm{i}}$ in the final sampling list increases (e.g., points A and C).
\item \textbf{Sorting \& Quantization:} Finally, we sort $\theta_{\mathrm{i}\in\{1,\dots,M\}}$ in descending order, and keep the $N$ points ($N \ll M$) corresponding to the $N$ highest $\theta_{\mathrm{i}}$. These $N$ points, which are then quantized (we use $1000$ quantization bins, by default) and serialized as in Equation \ref{eq:quantize}, denote the final sampled list.
\end{itemize}

The right column of Figure \ref{fig:sampling_illustration} depicts the adaptive sampling technique, which produces a better representation of sharp bends of the curve than uniform sampling, shown in the bottom-left of the same figure. We further illustrate the reconstruction from two techniques with a mask from the COCO dataset in Figure \ref{fig:sampling_example}, where the uniform sampling loses fine details of the zebra's legs, back, and ears. In contrast, adaptive sampling preserves the mask more precisely. 

Both uniform and adaptive sampling techniques inevitably result in a certain amount of information loss from the original ground-truth masks, thereby imposing a constraint on the maximal performance achievable in segmentation tasks. Nonetheless, the extent of this loss is considerably reduced when employing the adaptive sampling approach. For instance, in the RefCOCO validation set for Referring Expression Segmentation (RES), uniform sampling of $32$ points from the ground-truth masks yields an mIoU upper bound of $94.70$, whereas adaptive sampling achieves $97.26$. The superiority of adaptive sampling becomes even more pronounced in the case of complex geometric structures containing numerous sharp bends and intricate details. We delve deeper into the comparative efficacy of these two methods through ablation experiments in Section \ref{section:ablation}.


\vspace{-2.5mm}
\section{Coarse-to-fine Instruction-tuning Dataset}
\vspace{-2mm}
To train \model\ on a versatile form of vision and language tasks, we collect \data\ (\textbf{Co}arse-to-f\textbf{in}e \textbf{I}nstruction-\textbf{t}uning Dataset), which is a unified set of $14$ benchmarks containing $6.8M$ samples, among which $(i)$ $13$ are publicly available which we convert to instruction-tuning format, and ($ii$) we construct a new benchmark, AttCoSeg (\textbf{Att}ribute-level \textbf{Co}-\textbf{Seg}mentation), to alleviate the lack of multi-image region-level datasets. We quantitatively evaluate the trained model on $15$ benchmarks without additional fine-tuning. Notably, $4$ of these $15$ downstream contain entirely unseen tasks during training, helpful for assessing the system's generalization capability. To ensure data integrity, we confirm that no images from the validation or test sets appear during training, thus eliminating the risk of data leakage. We have grouped these diverse tasks into four main categories based on their input and output formats, summarized in Table \ref{tab:dataset_details}:

\begin{table}[!t]
\centering

\small
\setlength{\tabcolsep}{4pt}
\resizebox{\columnwidth}{!}{\begin{tabular}{@{} l c| c c c | c c c @{}}

\toprule
\multirow{2}{*}{\textbf{Dataset}} & \multicolumn{1}{c|}{\multirow{2}{*}{\textbf{Task}}} & \multirow{2}{*}{\textbf{Corpus}} & \multirow{2}{0.8cm}{\centering \bf Multi img?} & \multirow{2}{0.8cm}{\centering \bf Reg. level?} & \multirow{2}{1.0 cm}{\centering \bf Input format} & \multirow{2}{1.0 cm}{\centering \bf Output format} & \multirow{2}{*}{\textbf{Metrics (\%)}} \\ 

& & & & & & & \\

\midrule 

\multirow{8}{*}{COCO \cite{lin2014microsoft}} & Caption & Train, Eval & \ding{55} & \ding{55} & I & T & SPICE, CIDEr \\
& VQAv2 & Train, Eval & \ding{55} & \ding{55} & I + Q & T & Accuracy \\
& REC & Train, Eval & \ding{55} & \ding{51} & I + R & B & Pr@0.5 \\
& GREC & Train, Eval & \ding{55} & \ding{51} & I + R & M & Pr@0.5, N-acc\\
& RES & Train, Eval & \ding{55} & \ding{51} & I + R & B & mIoU \\
& GRES & Train, Eval & \ding{55} & \ding{51} & I + R & M & gIoU, N-acc, T-acc \\
& REG & Train & \ding{55} & \ding{51} & I + B & T & $-$ \\
& \CC{Light} AttCoSeg & \CC{Light} Train & \CC{Light} \ding{51} & \CC{Light} \ding{51} & \CC{Light} I & \CC{Light} M & \CC{Light} $-$ \\
\midrule
Flickr \cite{plummer2015flickr30k} & Spot Caption & Train & \ding{55} & \ding{51} & I & T + B & $-$ \\
VG \cite{krishna2017visual} & REG & Train & \ding{55} & \ding{51} & I + B & T & $-$ \\
VCR \cite{zellers2019recognition} & Reasoning & Train, Eval & \ding{55} & \ding{51} & I + Q + B & T & Accuracy\\ 
LLaVa \cite{liu2023llava} & VQA & Train & \ding{55} & \ding{55} & I + Q & T & $-$ \\
LT-QA \cite{mani2020point} & BQA & Train, Eval & \ding{55} & \ding{51} & I + Q + B & B & Accuracy\\
\multirow{2}{*}{V7W \cite{zhu2016visual7w}} & PQA & Train, Eval & \ding{55} & \ding{51} & I + Q + P & T & Accuracy\\
& BQA & Train, Eval & \ding{55} & \ding{51} & I + Q + B & T & Accuracy\\
TextVQA \cite{singh2019towards} & Reading comp. & Eval & \ding{55} & \ding{51} & I + Q & T & Accuracy\\
IconQA \cite{lu2021iconqa} & Reasoning & Eval & \ding{51} & \ding{51} & I + Q & T & Accuracy\\
HM \cite{kiela2020hateful} & Classification & Eval & \ding{55} & \ding{55} & I & T & Accuracy\\
POPE \cite{li2023evaluating} & Hallucination & Eval & \ding{55} & \ding{55} & I + Q & Y/N & Prec., Recall, F1\\

NLVR \cite{suhr2017corpus, suhr2019corpus} & Reasoning & Train, Eval & \ding{51} & \ding{55} & I + Q & Y/N & Accuracy\\
\midrule
PASCAL \cite{faktor2013co} & \multirow{3}{*}{CoSeg} & \multirow{3}{*}{Train, Eval} & \multirow{3}{*}{\ding{51}} & \multirow{3}{*}{\ding{51}} & \multirow{3}{*}{I} & \multirow{3}{*}{M} & \multirow{3}{2.25cm}{\centering Precision ($\mathcal{P}$), Jaccard Index ($\mathcal{J}$)} \\
iCoSeg \cite{batra2010icoseg} & & & & & \\
MSRC \cite{winn2005object} & & & & &\\

\bottomrule
\end{tabular}}
\vspace{-2mm}
\caption{\textbf{Training and evaluation datasets, input-output formats, and metrics.} To train \model\ on versatile form of vision and language tasks, we collect \data, which is a unified set of $14$ benchmarks. We quantitatively evaluate the trained model on $15$ tasks without additional fine-tuning, among which TextVQA, IconQA, POPE, and HM contain unseen tasks during training, assessing the system's generalization capability. I: Image, T: General Text, Q: Question, R: Referring Expression, P: Point coordinate, B: Bounding Box,  M: Segmentation Mask, Y/N: Yes or No.} \label{tab:dataset_details}
\vspace{-5.5mm}
\end{table}

\begin{itemize}[leftmargin=*]
\item Single-image coarse-level tasks, such as visual question answering (VQA) and image captioning on COCO \cite{lin2014microsoft} and LLaVa \cite{liu2023llava} require global understanding of a single input image. 
\item Single-image region-level tasks, like generalized referring expression comprehension (GREC) \cite{he2023grec} and segmentation (GRES) \cite{liu2023gres}, spot captioning \cite{chen2023shikra}, visual commonsense reasoning (VCR) \cite{zellers2019recognition}, box question answering (BQA) and point question answering (PQA) \cite{mani2020point, zhu2016visual7w} require fine-grained dense predictions over one input image. These tasks contain points, bounding boxes and segmentation masks in inputs and outputs.
\item Multi-image coarse-level tasks, like natural language for visual reasoning (NLVR) \cite{suhr2017corpus, suhr2019corpus} and icon question answering (IconQA) \cite{lu2021iconqa} involve comprehending global perception across multiple input images.
\item Multi-image region-level tasks, such as object-level co-segmentation (CoSeg) \cite{rother2006cosegmentation, li2022toward} demands fine-grained reasoning and grounding on various input images.  
\end{itemize}

\begin{table}[!t]
\centering

\small
\setlength{\tabcolsep}{4pt}
\resizebox{0.725\columnwidth}{!}{\begin{tabular}{l c | c c c c c}

\toprule

\multirow{2}{0.5cm}{\bf Method} & \multirow{2}{1.21cm}{\bf \centering General-purpose?} & \multicolumn{3}{c}{\bf VQAv2} & \multicolumn{2}{c}{\bf COCO Cap.}  \\ 

& & Val & Dev & Std & SPICE & CIDEr \\

\midrule

METER \cite{dou2022empirical} & \ding{55} & $-$ & 76.4 & 76.4 & 23.0 & \underline{128.2} \\
FIBER \cite{dou2022coarse} & \ding{55} & $-$ & \underline{78.6} & \underline{78.4} & \underline{23.1} & \bf 128.4 \\
Unified-IO \cite{lu2022unified} & \ding{51} & $-$ & 77.9 & $-$ & $-$ & 122.3 \\
Flamingo-80B \cite{alayrac2022flamingo} & \ding{51} & $-$ & 56.3 & $-$ & $-$ & 84.3 \\
Shikra-13B \cite{chen2023shikra} & \ding{51} & \underline{75.3} & 77.4 & 77.5 & $-$ & 117.5 \\
\rowcolor{Light}
\model-13B & \ding{51} & \bf 76.9 & \bf 79.1 & \bf 79.0 & \bf 23.3 & \bf 128.4 \\
\midrule

\bf \textcolor{blue}{$\Delta_{\text{Ours - Shikra-13B}}$} & $-$ & \textcolor{blue}{1.6} \textcolor{blue}{$\uparrow$} & \textcolor{blue}{1.7} \textcolor{blue}{$\uparrow$} & \textcolor{blue}{1.5} \textcolor{blue}{$\uparrow$} & $-$ & \textcolor{blue}{10.9} \textcolor{blue}{$\uparrow$} \\

\bottomrule
\end{tabular}}
\vspace{-2mm}
\caption{\textbf{Performance on VQAv2 and COCO captioning.} \model\ yields significant gains over existing general-purpose and fine-tuned baselines. Reported captioning results of METER and FIBER are without CIDEr optimization \cite{rennie2017self}.} \label{tab:vqa_caption}
\vspace{-6mm}
\end{table}

\noindent \textbf{AttCoSeg, newly proposed benchmark:} Existing multi-image region-level object co-segmentation datasets \cite{faktor2013co, batra2010icoseg, winn2005object} are small-scale and simple to solve. Hence, we argue that these datasets are insufficient to train \model\ to have generalized grounding ability over many input images, and we construct a more challenging larger-scale multi-image region-level dataset. We use Group-wise RES \cite{wu2023advancing} annotations to sample high-quality images containing objects with similar fine-grained attributes (shape, color, size, position). We refer to such images as positives. While training \model, we input these positive image pairs, ask the model to segment the object with common traits in both of them. We name this task attribute-level co-segmentation (AttCoSeg), which contains over 804k training samples, and help \model\ to gain significant generalized reasoning and grounding ability over multiple input images. Notably, we do not collect new images or perform new annotations ourselves when constructing AttCoSeg. Detailed statistics of every dataset are given in the supplementary.


\begin{table*}[!t]
\centering
\begin{subtable}[c]{0.48\textwidth}
\centering
\small
\setlength{\tabcolsep}{4pt}
\resizebox{\textwidth}{!}{\begin{tabular}{l c | c c c c c c c c }

\toprule 

\multirow{2}{*}{\bf Method} & \multirow{2}{1.21cm}{\bf \centering General-purpose?} & \multicolumn{3}{c}{Ref} & \multicolumn{3}{c}{Ref+} & \multicolumn{2}{c}{Refg} \\ 

& & val & testA & testB & val & testA & testB & val & test  \\

\midrule 

UniTAB \cite{yang2022unitab} & \ding{55} & 86.3 & 88.8 & 80.6 & 78.7 & 83.2 & 69.5 & 80.0 & 80.0 \\
MDETR \cite{kamath2021mdetr} & \ding{55} & 86.8 & 89.6 & 81.4 & 79.5 & 84.1 & 70.6 & 81.6 & 80.9 \\ 
SeqTR \cite{zhu2022seqtr} & \ding{55} & 83.7 & 86.5 & 81.2 & 71.5 & 76.3 & 64.9 & 74.9 & 74.2 \\
OFA-L \cite{wang2022ofa} & \ding{51} & 80.0 & 83.7 & 76.4 & 68.3 & 76.0 & 61.8 & 67.6 & 67.6 \\
VisionLLM-H \cite{wang2023visionllm} & \ding{51} & $-$ & 86.7 & $-$ & $-$ & $-$ & $-$ & $-$ & $-$ \\
Shikra-13B \cite{chen2023shikra} & \ding{51} & 87.8 & 91.1 & 81.8 & \underline{82.9} & 87.8 & 74.4 & 82.6 & 83.2 \\
MiniGPT-v2 \cite{chen2023minigptv2} & \ding{51} & 88.7 & 91.7 & \bf 85.3 & 80.0 & 85.1 & 74.5 & 84.4 & 84.7  \\
Ferret-13B \cite{you2023ferret} & \ding{51} & \underline{89.5} & \underline{92.4} & 84.4 & 82.8 & \underline{88.1} & \underline{75.2} & \underline{85.8} & \underline{86.3} \\
\rowcolor{Light}
\model-7B & \ding{51} & 88.1 & 91.5 & 83.0 & 82.9 & 89.8 & 74.8 & 83.6 & 84.4 \\
\rowcolor{Light}
\model-13B & \ding{51} & \bf 89.9 & \bf 92.5 & \underline{85.0} & \bf 84.1 & \bf 90.3 & \bf 75.8 & \bf 86.0 & \bf 86.4 \\

\midrule 

\bf \textcolor{blue}{$\Delta_{\text{Ours - Ferret-13B}}$} & $-$ & \textcolor{blue}{0.4} \textcolor{blue}{$\uparrow$} & \textcolor{blue}{0.1} \textcolor{blue}{$\uparrow$} & \textcolor{blue}{0.6} \textcolor{blue}{$\uparrow$} & \textcolor{blue}{1.3} \textcolor{blue}{$\uparrow$} & \textcolor{blue}{2.2} \textcolor{blue}{$\uparrow$} & \textcolor{blue}{0.6} \textcolor{blue}{$\uparrow$} & \textcolor{blue}{0.2} \textcolor{blue}{$\uparrow$} & \textcolor{blue}{0.1} \textcolor{blue}{$\uparrow$} \\

\bottomrule
\end{tabular}}
\subcaption{\textbf{Performance on referring expression comprehension (REC).} \model\ yields better results than existing baselines across all splits.}
\label{tab:rec}
\end{subtable}
\hspace{0.3em}
\begin{subtable}[c]{0.466\textwidth}
\centering
\small
\setlength{\tabcolsep}{4pt}
\resizebox{\textwidth}{!}{\begin{tabular}{l c | c c c c c c c c }

\toprule 

\multirow{2}{*}{\bf Method} & \multirow{2}{1.21cm}{\bf \centering General-purpose?} & \multicolumn{3}{c}{Ref} & \multicolumn{3}{c}{Ref+} & \multicolumn{2}{c}{Refg} \\ 

& & val & testA & testB & val & testA & testB & val & test  \\

\midrule 

CGAN \cite{luo2020cascade} & \ding{55} & 64.9 & 68.0 & 62.1 & 51.0 & 55.5 & 44.1 & 51.0 & 51.7 \\
VLT \cite{liu2022instance} & \ding{55} & 65.7 & 68.3 & 62.7 & 55.5 & 59.2 & 49.4 & 53.0 & 56.7\\ 
LTS \cite{jing2021locate} & \ding{55} & 65.4 & 67.8 & 63.1 & 54.2 & 58.3 & 48.0 & 54.4 & 54.3 \\
CRIS \cite{wang2022cris} & \ding{55} & 70.5 & 73.2 & 66.1 & 62.3 & 68.1 & 53.7 & 59.9 & 60.4 \\
SeqTR \cite{zhu2022seqtr} & \ding{55} & 71.7 & 73.3 & 69.8 & 63.0 & 66.7 & 59.0 & 64.7 & 65.7 \\
RefTr \cite{li2021referring} & \ding{55} & 74.3 & 76.8 & 70.9 & 66.8 & 70.6 & 59.4 & 66.6 & 67.4 \\
LAVT \cite{yang2022lavt} & \ding{55} & 74.5 & 76.9 & 70.9 & 65.8 & 71.0 & 59.2 & 63.3 & 63.6 \\
PolyFormer \cite{liu2023polyformer} & \ding{55} & \underline{76.0} & \underline{77.1} & \underline{73.2} & \underline{70.7} & \bf 74.5 & \underline{64.6} & \underline{69.4} & \underline{69.9} \\

\rowcolor{Light}
\model-7B & \ding{51} & 74.5 & 76.0 & 72.7 & 69.1 & 73.7 & 64.0 & 69.0 & 70.9 \\
\rowcolor{Light}
\model-13B & \ding{51} & \bf 77.2 & \bf 78.7 & \bf 73.9 & \bf 71.8 & \underline{74.4} & \textbf{65.6} & \bf 69.8 & \bf 71.9 \\

\midrule 

\bf \textcolor{blue}{$\Delta_{\text{Ours - PolyFormer}}$} & $-$ & \textcolor{blue}{1.2} \textcolor{blue}{$\uparrow$} & \textcolor{blue}{1.6} \textcolor{blue}{$\uparrow$} & \textcolor{blue}{0.7} \textcolor{blue}{$\uparrow$} & \textcolor{blue}{1.1} \textcolor{blue}{$\uparrow$} & \textcolor{Brown}{0.1} \textcolor{Brown}{$\downarrow$} & \textcolor{blue}{1.0} \textcolor{blue}{$\uparrow$} & \textcolor{blue}{0.4} \textcolor{blue}{$\uparrow$} & \textcolor{blue}{2.0} \textcolor{blue}{$\uparrow$} \\

\bottomrule
\end{tabular}}
\subcaption{\textbf{Performance on referring expression segmentation (RES).} \model\ is the first general-purpose model to unify RES.}
\label{tab:res}
\end{subtable}
\vspace{-3mm}
\caption{\textbf{Performance on (a) REC, and (b) RES.} While none other general-purpose systems can solve RES, \model\ sets a new state-of-the-art for both tasks across all splits.}
\label{tab:rec_res}
\vspace{-5mm}
\end{table*}

\vspace{-2mm}
\section{Experiments} 
\vspace{-1mm}

\subsection{Instruction Prompts}
\vspace{-1mm}
Carefully designed language instructions are crucial for general-purpose vision models on diverse tasks with different input-output formats \cite{wang2023visionllm, chen2023shikra}. Since we address closely related tasks like REC, RES, GREC, GRES, we use detailed instructions. Figure \ref{fig:system_main} illustrates an example instruction for CoSeg. More example instructions are shown in supplementary. We use a special token $<$image$>$, which we later replace with the instruction-guided image features to generate interleaved image-text input to the LLM.   


Moreover, the instruction must vary for different samples to support flexible user inputs. To generate high-quality instructions with minimal cost, we manually write one example description of each task and resort to GPT-3.5 \cite{brown2020language} to create hundreds of variations. Next, we refine and ensure the quality of every instruction with GPT-4 \cite{OpenAI2023GPT4TR}. During training, we randomly pick one instruction for each sample. 


\begin{table}[!t]

\centering

\small
\setlength{\tabcolsep}{4pt}
\resizebox{\columnwidth}{!}{\begin{tabular}{l c | c c | l c | c c c}

\toprule

\multirow{2}{0.5cm}{\bf Method} & \multirow{2}{1.21cm}{\bf \centering General-purpose?} & \multicolumn{2}{c|}{\bf GREC}  & \multirow{2}{*}{\bf Method} & \multirow{2}{1.21cm}{\bf \centering General-purpose?} & \multicolumn{3}{c}{\bf GRES}  \\ 

& & Pr & N-acc. & & & gIoU & N-acc. & T-acc. \\

\midrule

MCN \cite{luo2020mcn} & \ding{55} & 28.0 & 30.6 & MattNet \cite{yu2018mattnet} & \ding{55} & 48.2 & 41.2 & 96.1 \\
VLT \cite{liu2022instance} & \ding{55} & 36.6 & 35.2 & VLT \cite{liu2022instance} & \ding{55} & 52.0 & 47.2 & 95.7  \\
MDETR \cite{kamath2021mdetr} & \ding{55} & \underline{41.5} & \underline{36.1} & LAVT \cite{yang2022lavt} & \ding{55} & \underline{58.4} & \underline{49.3} & \underline{96.2} \\

\rowcolor{Light}

\model-7B & \ding{51} & 52.7 & 69.4 & \model-7B & \ding{51} & 64.4 & 68.8 & 96.6 \\
\rowcolor{Light}

\model-13B & \ding{51} & \bf 54.6 & \bf 70.8 & \model-13B & \ding{51} & \bf 65.1 & \bf 70.0 & \bf 96.8 \\

\midrule 

\bf \textcolor{blue}{$\Delta_{\text{Ours - MDETR}}$} & $-$ & \textcolor{blue}{13.1} \textcolor{blue}{$\uparrow$} & \textcolor{blue}{34.7} \textcolor{blue}{$\uparrow$} & \bf \textcolor{blue}{$\Delta_{\text{Ours - LAVT}}$} & $-$ & \textcolor{blue}{6.7} \textcolor{blue}{$\uparrow$} & \textcolor{blue}{20.7} \textcolor{blue}{$\uparrow$} & \textcolor{blue}{0.6} \textcolor{blue}{$\uparrow$}  \\

\bottomrule
\end{tabular}}
\vspace{-1.5mm}
\caption{\textbf{Performance on generalized referring expression comprehension (GREC) and generalized referring expression segmentation (GRES).} \model\ is the first general-purpose system to address both tasks, and gains huge improvements over existing specialist models.} \label{tab:grec_gres}
\vspace{-6mm}
\end{table}

\vspace{-1mm}
\subsection{Implementation Details} \label{sec:implementation_details}
\vspace{-1mm}

We use EVA-CLIP \cite{sun2023eva} pre-trained on LAION-$400$M \cite{schuhmann2021laion} and QFormer \cite{li2023blip2} pre-trained by InstructBLIP \cite{instructblip} as our visual encoder and instruction-guided image tokenizer. We feed the input images into EVA, which produces $256 \times 1408$ dimensional features for $224 \times 224$ images. The number of spatial tokens quadratically increases with the input image dimension. The Qformer has $12$ encoder layers with $12$ heads and outputs $32$ queries per image with a hidden size of $768$, thus working as an efficient feature compressor. For a fair comparison with existing general-purpose baselines \cite{wang2023visionllm, chen2023shikra, you2023ferret, zhang2023gpt4roi, chen2023minigptv2}, we use Vicuna$7$B and Vicuna$13$B \cite{chiang2023vicuna} as the LLM. All other dense layers are initialized from scratch. For serializing the segmentation masks, we sample $32$ points using the proposed adaptive sampling technique.

\model\ is trained in two stages. In the first stage, we only use the single-image datasets and do not introduce the instruction-guided image tokenizer. We freeze EVA and train the rest of the model end-to-end for $2$ epochs. In the second stage, we only tune the image tokenizer on the multi-image datasets for $5$ epochs. \model\ is trained using AdamW optimizer \cite{loshchilov2018decoupled} and cosine scheduler \cite{loshchilov2016sgdr} with linear warmup for the first $3$\% steps. We use a peak learning rate of $2$e$-5$ and a global batch size of $256$. The model from the first stage is used to evaluate single-image datasets, whereas the model from the second stage is used to evaluate multi-image datasets. Training takes ~$2/3$ days for the first stage and ~$22/30$ hours for the second stage with $7/13$B models on $32$ A$100$ GPUs, each having $80$G memory.

\begin{table}[!t]
\centering
\small
\setlength{\tabcolsep}{4pt}
\resizebox{0.97\columnwidth}{!}{\begin{tabular}{@{} l l | c c c | l l | c @{}}

\toprule

\multirow{2}{*}{\bf \centering Task} & \multirow{2}{*}{\bf \centering Method} & \multicolumn{3}{c|}{\bf \centering LookTwice-QA} & \multirow{2}{*}{\bf \centering Task} & \multirow{2}{*}{\bf \centering Method} & \multirow{2}{*}{\bf \centering V7W} \\

& & Any & Super cls. & Object & &  \\

\midrule

\multirow{4}{*}{PQA} & Mani et al. \cite{mani2020point} & 56.5 & 59.1 & 62.8 & \multirow{7}{*}{BQA} & V7W \cite{zhu2016visual7w} & 56.1 \\
& Shikra-13B \cite{chen2023shikra} & \underline{70.0} & \underline{70.2} & \underline{71.9} & & CMNs \cite{hu2017modeling} & 72.5 \\
& \CC{}\model-13B & \CC{}\bf 71.1 & \CC{}\bf71.2 & \CC{}\bf72.5 & & ViLBERT \cite{lu202012} & 82.8 \\

& \bf \textcolor{blue}{$\Delta_{\text{Ours - Shikra-13B}}$} & \textcolor{blue}{1.1} \textcolor{blue}{$\uparrow$} & \textcolor{blue}{1.0} \textcolor{blue}{$\uparrow$} & \textcolor{blue}{0.6} \textcolor{blue}{$\uparrow$} & & ViLBERT$_{\text{FT}}$ \cite{lu202012} & 83.4 \\

\multirow{3}{*}{BQA} & Mani et al. \cite{mani2020point} & 60.2 & 59.8 & 61.4 & & GPT4RoI-13B \cite{zhang2023gpt4roi} & 84.8 \\
& Shikra-13B \cite{chen2023shikra} & \underline{70.3} & \underline{71.4} & \underline{72.3} & & Shikra-13B \cite{chen2023shikra} & \underline{85.3} \\
& \CC{}\model-13B & \CC{}\bf71.4 & \CC{}\bf72.5 & \CC{}\bf73.0 & & \CC{}\model-13B & \CC{}\bf85.5 \\
\midrule
& \bf \textcolor{blue}{$\Delta_{\text{Ours - Shikra-13B}}$} & \textcolor{blue}{1.1} \textcolor{blue}{$\uparrow$} & \textcolor{blue}{1.1} \textcolor{blue}{$\uparrow$} & \textcolor{blue}{0.7} \textcolor{blue}{$\uparrow$} & & \bf \textcolor{blue}{$\Delta_{\text{Ours - Shikra-13B}}$} & \textcolor{blue}{0.2} \textcolor{blue}{$\uparrow$} \\

\bottomrule

\end{tabular}}
\vspace{-2mm}
\caption{\textbf{Performance of point question answering (PQA) and box question answering (BQA) on LookTwice-QA and Visual-7W.} LookTwice-QA questions based on input point/box on three different level of referential clarity in the question, e.g. ``How many of these [items/vehicles/cars] are there?" Visual-7W questions in 'which box' setting, i.e. choose one of the four bounding box options based on given query.}\label{tab:pqa_bqa}
\vspace{-5.5mm}
\end{table}

\vspace{-2mm}
\subsection{Main Results}
\vspace{-1mm}

We use \textbf{boldface} and \underline{underline} for the best and second-best performing methods in every table and indicate the performance improvements over the state-of-the-art with \textcolor{blue}{$\Delta$}.

\vspace{1mm}

\noindent \textbf{VQAv2 \& COCO Captioning:} Table \ref{tab:vqa_caption} presents the performance on traditional single-image coarse-level visual question answering and image captioning tasks, which do not necessitate coordinates in the input or output. The input instructions for these tasks are straightforward, such as, ``Please generate a simple description of the image $<$image$>$." or ``Given the image $<$image$>$, can you please answer the question $<$question$>$", where $<$question$>$ denotes the input query. On VQAv2, \model\ achieves $76.9\%$, $79.1\%$, and $79.0\%$ accuracy on the val, dev, and std splits, improving the general-purpose state-of-the-art by over $1.5$ points. On image captioning, \model\ yields a substantial gain of $10.9$ CIDEr points over the best general-purpose baseline \cite{chen2023shikra}. Our model performs on a par with fine-tuned specialist models, signifying the power of LLMs to comprehend and generate strong language descriptions. 

\begin{table}
\begin{subtable}[c]{0.503\columnwidth}
\centering
\small
\setlength{\tabcolsep}{4pt}
\resizebox{\textwidth}{!}{\begin{tabular}{l | c c c}

\toprule 

\multirow{2}{*}{\bf Method} & \multicolumn{3}{c}{Validation Acc.} \\ 

& Q $\to$ A & QA $\to$ R & Q $\to$ AR \\

\midrule 

ViLBERT \cite{lu2019vilbert} & 72.4 & 74.5 & 54.0 \\
Unicoder-VL \cite{li2020unicoder} & 72.6 & 74.5 & 54.5 \\
VLBERT \cite{su2019vl} &  75.5 & 77.9 & 58.9\\
VILLA \cite{gan2020large} & 78.5 & 82.6 & 65.2 \\
GPT4RoI-7B \cite{zhang2023gpt4roi} & \underline{87.4} & \underline{89.6} & \underline{78.6} \\

\rowcolor{Light}
\model-13B & \bf 87.8 & \bf 89.9 & \bf 79.1 \\
\midrule
\bf \textcolor{blue}{$\Delta_{\text{Ours - GPT4RoI-7B}}$} & \textcolor{blue}{0.4} \textcolor{blue}{$\uparrow$} & \textcolor{blue}{0.3} \textcolor{blue}{$\uparrow$} & \textcolor{blue}{0.5} \textcolor{blue}{$\uparrow$} \\

\bottomrule
\end{tabular}}
\subcaption{\textbf{Performance on visual commonsense reasoning (VCR).}}
\label{tab:vcr}
\end{subtable}
\hspace{0.125em}
\begin{subtable}[c]{0.46\columnwidth}
\centering
\small
\setlength{\tabcolsep}{4pt}
\resizebox{\textwidth}{!}{\begin{tabular}{l | c c c}

\toprule 

\multirow{2}{*}{\bf Method} & \multicolumn{3}{c}{Acc.} \\ 

& TextVQA & IconQA & HM \\

\midrule 

BLIP-2 \cite{li2023blip2} & 42.5 & 40.6 & 53.7 \\
InstructBLIP \cite{instructblip} & 50.7 & 44.8 & 57.5 \\
MiniGPT-4 \cite{zhu2023minigpt} & 19.9 & 37.6 & $-$ \\
LLaVA \cite{liu2023llava} & 38.9 & 43.0 & $-$ \\
MiniGPT-v2 \cite{chen2023minigptv2} & \underline{51.9} & \underline{47.7} & \underline{58.2} \\

\rowcolor{Light}
\model-13B & \bf 53.0 & \bf 47.9 & \bf 59.1 \\
\midrule
\bf \textcolor{blue}{$\Delta_{\text{Ours - MiniGPTv2}}$} & \textcolor{blue}{1.1} \textcolor{blue}{$\uparrow$} & \textcolor{blue}{0.2} \textcolor{blue}{$\uparrow$} & \textcolor{blue}{0.9} \textcolor{blue}{$\uparrow$} \\

\bottomrule
\end{tabular}}
\subcaption{\textbf{Performance on novel tasks - TextVQA, IconQA, and HM. }}
\label{tab:novel}
\end{subtable}
\vspace{-3mm}
\caption{\textbf{Results on (a) VCR, and (b) three novel tasks - TextVQA, IconQA, hateful memes (HM).} \model\ achieves consistent gains over existing baselines.}
\label{tab:vcr_novel}
\vspace{-3mm}
\end{table}

\begin{table}[!t]
\centering
\small
\setlength{\tabcolsep}{4pt}
\resizebox{\columnwidth}{!}{\begin{tabular}{@{} l | c c | l | c c | l | c @{}}

\toprule

\multirow{2}{*}{\bf \centering Method} & \multicolumn{2}{c|}{\bf \centering PASCAL} & \multirow{2}{*}{\bf \centering Method} & \multicolumn{2}{c|}{\bf \centering MSRC} & \
\multirow{2}{*}{\bf \centering Method} & \multicolumn{1}{c}{\bf \centering iCoSeg} \\

& Av. $\mathcal{P}$ & Av. $\mathcal{J}$ & & Av. $\mathcal{P}$ & Av. $\mathcal{J}$ & & Av. $\mathcal{J}$  \\

\midrule

Quan et al. \cite{quan2016object} & 89.0 & 52.0 & Rubinstein et al. \cite{rubinstein2013unsupervised} & 92.2 & 74.7 & Rubinstein et al. \cite{rubinstein2013unsupervised} & 70.2 \\
Jerripothula et al. \cite{jerripothula2017object} & 80.1 & 40.0 & Faktor et al. \cite{faktor2013co} & 92.0 & 77.0 & Faktor et al. \cite{faktor2013co} & 73.8 \\
Li et al. \cite{li2019group} & 94.1 & 63.0 & Chen et al. \cite{chen2018semantic} & $-$ & 73.9 & Jerripothula et al. \cite{jerripothula2016image} & 70.4 \\
Zhang et al. \cite{zhang2020deep} & 94.9 & 71.0 & Li et al. \cite{li2019deep} & 95.4 & 82.9 & Zhang et al. \cite{zhang2020deep} & 89.2 \\
CycleSegNet \cite{zhang2021cyclesegnet} & \underline{96.8} & \underline{73.6} & CycleSegNet \cite{zhang2021cyclesegnet} & \underline{97.9} & \underline{87.2} & CycleSegNet \cite{zhang2021cyclesegnet} & \underline{92.1} \\

\rowcolor{Light}
\model-13B & \bf 97.9 & \bf 77.2 & \model-13B & \bf 98.5 & \bf 90.1 & \model-13B & \bf 95.1 \\

\midrule

\bf \textcolor{blue}{$\Delta_{\text{Ours - CycleSegNet}}$} & \textcolor{blue}{1.1} \textcolor{blue}{$\uparrow$} & \textcolor{blue}{3.6} \textcolor{blue}{$\uparrow$} & \textcolor{blue}{$\Delta_{\text{Ours - CycleSegNet}}$} & \textcolor{blue}{0.6} \textcolor{blue}{$\uparrow$} & \textcolor{blue}{2.9} \textcolor{blue}{$\uparrow$} & \textcolor{blue}{$\Delta_{\text{Ours - CycleSegNet}}$} & \textcolor{blue}{3.0} \textcolor{blue}{$\uparrow$} \\

\bottomrule

\end{tabular}}
\vspace{-2mm}
\caption{\textbf{Performance on object co-segmentation (CoSeg) on three datasets - PASCAL, MSRC, and iCoSeg.} \model\ is the first general-purpose system to address CoSeg and sets a new set-of-the-art across all datasets, beating previous specialist models.}\label{tab:coseg}
\vspace{-6mm}
\end{table}

\vspace{1mm}

\noindent \textbf{REC, RES, GREC \& GRES:} Next, we evaluate \model\ on four single-image grounding tasks. Table \ref{tab:rec_res} shows the results of referring expression comprehension (REC) and referring expression segmentation (RES), which aims to ground (detect and segment, respectively) one object in the image described by an input expression. Our model shows promising performance on REC, improving over existing baselines across all evaluation splits. \model\ is the first general-purpose system to report results on RES, where we perform as good as fine-tuned specialist models. Such strong results on grounding tasks can be attributed to refined image features, effective sampling techniques, and detailed input instructions. We also evaluate \model\ on GREC \& GRES, where the output can contain zero, one, or multiple boxes and masks. As shown in Table \ref{tab:grec_gres}, besides generating high-quality boxes and masks, our model yields an impressive gain of $34.7\%$ and $20.7\%$ N-acc scores over MDETR \cite{kamath2021mdetr}, reflecting the ability of \model\ to detect samples without any matching objects in the image. 


\vspace{1mm}

\noindent \textbf{PQA \& BQA:} Table \ref{tab:pqa_bqa} shows our performance on point question answering (PQA) and box question answering (BQA), which can have coordinate points and bounding boxes as input and output. LookTwice-QA asks the model to answer a question about a specified region, either mentioning a point or a box. The system needs to comprehend the area in the context of the whole image, e.g., ``How many of these [cars] are there in the image?" Visual-7W contains MCQs where the model needs to choose a box from four options. \model\ sets new state-of-the-art on both tasks, proving its mighty region-referring ability.

\vspace{1mm}
\noindent \textbf{VCR \& Novel (Unseen) Tasks:} Table \ref{tab:vcr} shows results on visual commonsense reasoning (VCR) - a single-image fine-grained reasoning task containing questions with referring bounding boxes. \model\ produces $0.5\%$ improvement over GPT4RoI \cite{zhang2023gpt4roi} in the most challenging $Q \to AR$ setting. We also access our model's generalization ability by evaluating it on three novel tasks in Table \ref{tab:novel} - TextVQA, IconQA, and hateful memes (HM). \model\ achieves strong results on all three benchmarks, proving its ability to comprehend novel tasks given well-designed instructions. 


\begin{table}
\centering
\begin{subtable}[c]{0.5135\columnwidth}
\centering
\small
\setlength{\tabcolsep}{4pt}
\resizebox{\textwidth}{!}{\begin{tabular}{l c | c c}

\toprule

\multirow{2}{0.5cm}{\bf Method} & \multirow{2}{1.21cm}{\bf \centering General-purpose?} & \multicolumn{2}{c}{\bf NLVR}  \\ 

& & dev & test-P \\

\midrule

VisualBERT \cite{li2019visualbert} & \ding{55} & 67.4 & 67.0 \\
SOHO \cite{huang2021seeing} & \ding{55} & 76.3 & 77.3\\
Oscar \cite{li2020oscar} & \ding{55} & 78.1 & 78.4 \\
Uniter \cite{chen2020uniter} & \ding{55} & 77.2 & 77.9 \\
VILLA \cite{gan2020large} & \ding{55} & 78.4 & 79.3 \\
ALBEF \cite{li2021align} & \ding{55} & \underline{80.2} & \underline{80.5} \\
\rowcolor{Light}
\model-13B & \ding{51} & \bf{80.8} & \bf{81.3} \\

\midrule

\bf \textcolor{blue}{$\Delta_{\text{Ours - ALBEF}}$} & $-$ & \textcolor{blue}{0.6} \textcolor{blue}{$\uparrow$} & \textcolor{blue}{0.8} \textcolor{blue}{$\uparrow$} \\

\bottomrule
\end{tabular}}
\subcaption{\textbf{Results on NLVR.}}
\label{tab:nlvr}
\end{subtable}
\hspace{0.1 em}
\begin{subtable}[c]{0.461\columnwidth}
\centering
\small
\setlength{\tabcolsep}{4pt}
\resizebox{\textwidth}{!}{\begin{tabular}{l | c c c}

\toprule

\multirow{2}{0.5cm}{\bf Method} & \multicolumn{1}{c}{R} & \multicolumn{1}{c}{P} & \multicolumn{1}{c}{A}  \\ 

& F1 & F1 & F1 \\

\midrule

mPLUG-Owl & 68.4 & 66.9 & 66.8 \\
LLaVA \cite{liu2023llava} & 66.6 & 66.4 & 66.3 \\
MiniGPT4 \cite{zhu2023minigpt} & 80.2 & 73.0 & 70.4 \\
InstructBLIP \cite{instructblip} & 89.3 & \underline{84.7} & 77.3 \\
Shikra-7B \cite{chen2023shikra} & 86.2 & 83.2 & \underline{82.5} \\
Ferret-13B \cite{you2023ferret} & \underline{89.8} & 84.2 & 82.0 \\

\rowcolor{Light}
\model-13B & \bf 90.5 & \bf 84.8 & \bf 82.9 \\
\midrule
\bf \textcolor{blue}{$\Delta_{\text{Ours - Ferret-13B}}$} & \textcolor{blue}{0.7} \textcolor{blue}{$\uparrow$} & \textcolor{blue}{0.6} \textcolor{blue}{$\uparrow$} & \textcolor{blue}{0.9} \textcolor{blue}{$\uparrow$} \\

\bottomrule
\end{tabular}}
\subcaption{\textbf{Results on POPE.}}
\label{tab:pope}
\end{subtable}
\vspace{-2mm}
\caption{\textbf{Performance on (a) NLVR, and (b) object hallucination benchmark using POPE evaluation pipeline}. \model\ is the first general-purpose model to address NLVR, and beats strong fine-tuned models. \model\ demonstrates an intriguing property of alleviating object hallucinations across all three splits. R: Random, P: Popular, A: Adversarial.}
\label{tab:nlvr_pope}
\vspace{-4mm}
\end{table}

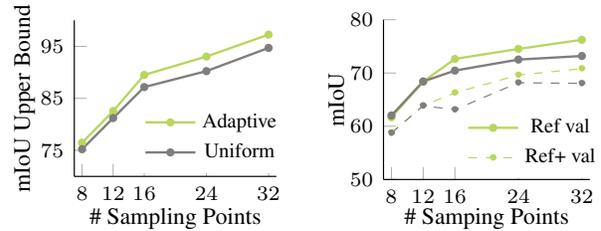
\begin{figure}[!t]
    \centering
    \hspace{-0.em}
    \begin{subfigure}[b]{0.481\linewidth}
        \resizebox{1.0\textwidth}{!}{
            \begin{tikzpicture}
	\begin{axis} [
		axis x line*=bottom,
		axis y line*=left,
		legend pos=north east,
		ymin=70, ymax=100,
		xmin=7, xmax=33,
		xticklabel={\pgfmathparse{\tick}\pgfmathprintnumber{\pgfmathresult}},
		xtick={8,12,16,24,32},
		ytick={75,85,95},
        xlabel={\footnotesize{\# Sampling Points}},
        ylabel={\footnotesize{mIoU Upper Bound}},
        xlabel shift=-4pt,
        ylabel shift=-5pt,
		width=\linewidth,
		legend style={cells={align=left}},
		label style={font=\scriptsize},
		tick label style={font=\scriptsize},
		legend style={draw=none,at={(0.30,0.25)},anchor=west},
		]

        \addplot[mark=*,mark options={scale=0.5, fill=Dark},style={thick},Dark] plot coordinates {
			(8, 76.47)
            (12, 82.55)
			(16, 89.51)
			(24, 93.04)
            (32, 97.26)
		};
        \addlegendentry{\scriptsize{Adaptive}}

        \addplot[mark=*,mark options={scale=0.5, fill=Salmon},style={thick},Salmon] plot coordinates {
			(8, 75.23)
            (12, 81.21)
			(16, 87.17)
			(24, 90.24)
            (32, 94.70)
		};
        \addlegendentry{\scriptsize{Uniform}}

	\end{axis}
\end{tikzpicture}%
        }
        \caption{\textbf{mIoU upper bound} on Ref val set with varying number of points.}
        \label{fig:sampling_ablation_1}
    \end{subfigure}
 \hspace{-0em}
    \begin{subfigure}[b]{0.488\linewidth}
        \resizebox{1.0\textwidth}{!}{
            \begin{tikzpicture}
	\begin{axis} [
		axis x line*=bottom,
		axis y line*=left,
		legend pos=north east,
		ymin=50, ymax=80,
		xmin=7, xmax=33,
		xticklabel={\pgfmathparse{\tick}\pgfmathprintnumber{\pgfmathresult}},
		xtick={8, 12, 16, 24, 32},
		ytick={50,60,70,80},
        xlabel={\footnotesize{\# Samping Points}},
        ylabel={\footnotesize{mIoU}},
        xlabel shift=-5pt,
        ylabel shift=-5pt,
		width=\linewidth,
		legend style={cells={align=left}},
		label style={font=\scriptsize},
		tick label style={font=\scriptsize},
		legend style={draw=none,at={(0.36,0.23)},anchor=west},
		]

        \addplot[mark=*,mark options={scale=0.5, fill=Dark},style={thick},Dark] plot coordinates {
			(8, 61.68)
            (12, 68.38)
            (16, 72.64)
            (24, 74.52)
			(32, 76.23)
		};
        \addlegendentry{\scriptsize{Ref val}}

        \addplot[mark=*,mark options={scale=0.5, fill=Dark},style={dashed},Dark] plot coordinates {
			(8, 58.78)
            (12, 63.86)
            (16, 66.29)
            (24, 69.61)
			(32, 70.82)
		};  \addlegendentry{\scriptsize{Ref+ val}}

		\addplot[mark=*,mark options={scale=0.5, fill=Salmon},style={thick},Salmon] plot coordinates {
			(8, 62.02)
            (12, 68.41)
            (16, 70.45)
            (24, 72.52)
			(32, 73.19)
		};  

        \addplot[mark=*,mark options={scale=0.5, fill=Salmon},style={dashed},Salmon] plot coordinates {
			(8, 58.73)
            (12, 63.84)
            (16, 63.12)
            (24, 68.17)
			(32, 68.05)
		};

	\end{axis}
\end{tikzpicture}%
        }
        \caption{\textbf{mIoU by \model} on Ref, Ref+ with varying number of points.}
        \label{fig:sampling_ablation_2}
    \end{subfigure}
\vspace{-6mm}
\caption{\textbf{Ablative experiments on RES task.} (a) Comparison of the highest possible mIoU by adaptive and uniform sampling, indicating lesser information loss in adaptive sampling, (b) Effect of number of sampled points on the performance of \model.}
\label{fig:sampling_ablation}
\vspace{-6mm}
\end{figure}


\begin{figure*}[!t]
\centering
\includegraphics[width=0.97\textwidth]{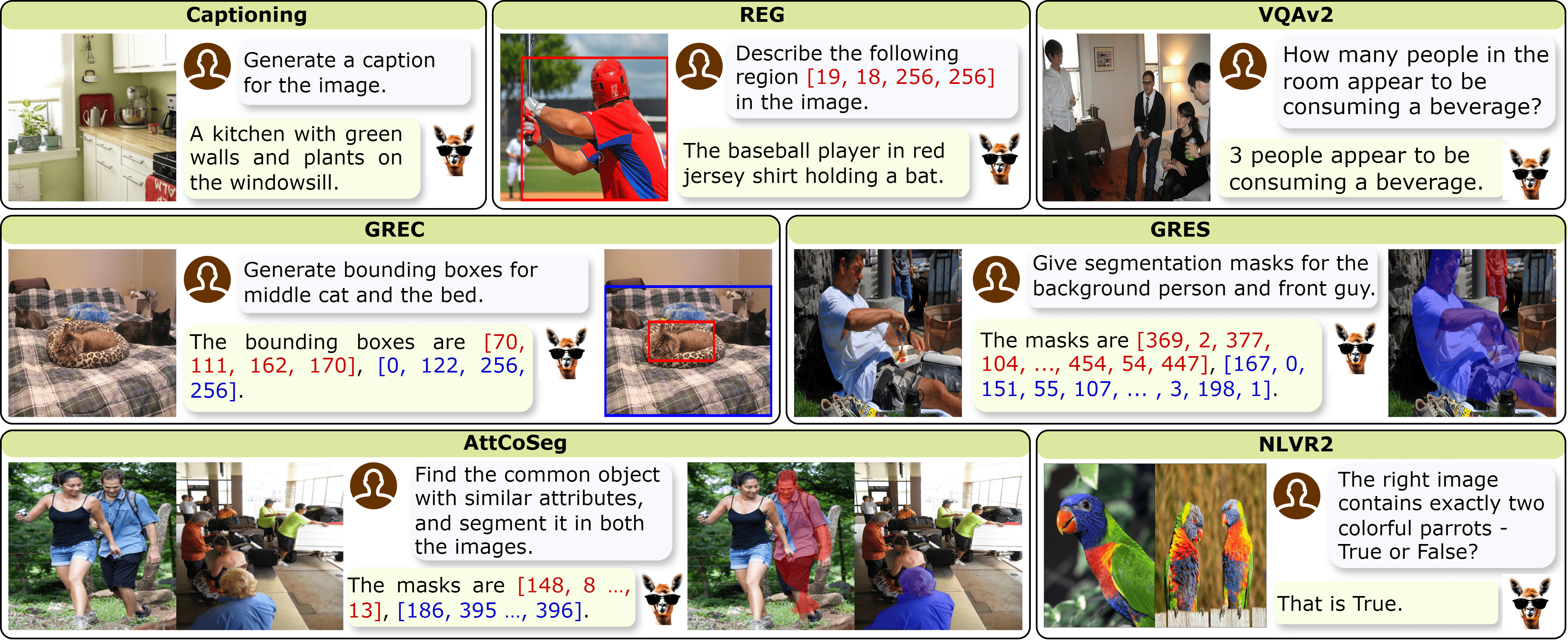}
\vspace{-3mm}
\caption{\textbf{Examples demonstrating \model's capability for single and multi-image reasoning and grounding tasks.} More visualizations are shown in supplementary. Best
viewed when zoomed in and in color.}
\vspace{-4mm}
\label{fig:visualization_main}
\end{figure*}

\vspace{1mm}

\noindent \textbf{CoSeg \& NLVR:} Table \ref{tab:coseg} and Table \ref{tab:nlvr} shows the performance on two multi-image tasks, CoSeg and NLVR. \model\ is the first general-purpose model to evaluate both tasks. Given a group of images with a common object, CoSeg aims to recognize and segment the object in every photo. \model\ outperforms existing specialist baselines across three different datasets on CoSeg, showing its strong perception and grounding ability. \model\ also beats powerful fine-tuned models \cite{gan2020large, li2021align} on NLVR, which aim to reason two input images and answer a query. These results prove the versatility of \model\ with more than one input image, which is crucial for real-world use cases.

\vspace{1mm}

\noindent \textbf{POPE:} We evaluate \model\ on POPE object hallucination benchmark in Table \ref{tab:pope}, where we perform comparably to strong general-purpose models like Shikra \cite{chen2023shikra}, and Ferret \cite{you2023ferret} across all metrics and splits, and vastly outperform many previous baselines. These results exhibit our model's ability to power against the hallucination problem, essential for its generalized applicability. 











\begin{SCtable}[][!t]
\vspace{10mm}
\centering

\small
\setlength{\tabcolsep}{4pt}
\resizebox{0.5\columnwidth}{!}{\begin{tabular}{l | c c c c c c c}

\toprule 

\multirow{2}{0.5cm}{\bf Method} & \multicolumn{1}{c}{\bf iCoSeg} & \multicolumn{1}{c}{\bf NLVR} \\ 

& Av. $\mathcal{J}$ & dev \\

\midrule

\rowcolor{Light}
\model-13B & \bf 95.1 & \bf 80.8 \\
w/o Tokenizer & 89.7 & 77.3  \\
w/o Tokenizer PT & 94.8 & 79.5 \\

\bottomrule
\end{tabular}}
\vspace{-12mm}
\caption{\textbf{Ablation on instruction-guided image tokenizer}, which refines global image embeddings.} \label{tab:qformer_ablation}
\vspace{-20mm}
\end{SCtable}

\vspace{-2mm}
\subsection{Ablation Study} \label{section:ablation}
\vspace{-1mm}
\noindent \textbf{Adaptive vs. Uniform Sampling:} We ablate the quantitative effectiveness of our proposed adaptive sampling method compared to uniform sampling for referring expression segmentation (RES) in Figure \ref{fig:sampling_ablation}. With $32$ sampled points, the maximum achievable mIoU score on Ref val set by adaptive technique is $97.26$, while for uniform sampling, $94.70$. However, with fewer sampling points, both methods perform significantly worse. Figure \ref{fig:sampling_ablation_2} shows that the performance of \model\ also improves using adaptive sampling on both Ref and Ref+ val splits, which shows the usefulness of the proposed sampling scheme.

\noindent \textbf{Number of Sampled Points:} Figure \ref{fig:sampling_ablation_2} shows that with a higher number of sampled points, the performance of \model\ significantly improves for both Ref and Ref+. When increasing the number of points from $16$ to $32$, \model\ gains $3.6$ on Ref and $4.5$ on Ref+. 


\noindent \textbf{Instruction-guided Image Tokenizer:} We ablate the importance of the proposed instruction-guided tokenizer in Table \ref{tab:qformer_ablation}. The performance of iCoSeg significantly drops by $5.4$ $\mathcal{J}$-index without the tokenizer module. We also see similar effects in captioning, RES, VCR, and NLVR. When using QFormer without pre-trained weights, we observe a substantial drop in all tasks except iCoSeg.    

\noindent \textbf{LLM Size:} Table \ref{tab:vqa_caption}, \ref{tab:rec_res}, \ref{tab:grec_gres} shows that larger LLM backbone generally helps improve the performance. We show ablation on the training dataset and image encoder in supplementary. 

\vspace{-2mm}
\subsection{Qualitative Results and Error Analysis}
\vspace{-1mm}
Figure \ref{fig:visualization_main} visualizes sample results from \model\ for single and multi-image reasoning and grounding tasks. As shown in the NLVR and AttCoSeg examples, \model\ can successfully parse all input images and comprehend the relation among them. It can also successfully ground all referred objects in foreground and background, as shown in GRES. However, compared to the recently released GPT-4V \cite{yang2023dawn}, we perform worse in general and knowledge-based question answering, which can be attributed to the billion scale pre-training of GPT. Nevertheless, \model's ability to reason over several images and perform precise detection and segmentation makes it unique.

\vspace{-2mm}
\section{Conclusion}
\vspace{-1mm}

We introduce \model, a powerful general-purpose vision system that integrates coarse- and fine-grained vision-language reasoning and grounding tasks over single and multiple input images into a unified framework. To filter embeddings from various images, \model\ uses a language-guided image tokenizer, which provides compressed and refined features following the task description. We also employ a gradient-aware adaptive sampling technique to efficiently represent binary segmentation masks as sequences, significantly improving previously used uniform sampling. We conduct extensive experiments to show the effectiveness of \model\ on a wide range of downstream tasks, consistently achieving state-of-the-art performance. 

\vspace{-2mm}
\section{Acknowledgement}
\vspace{-1mm}

The codebase for this work is built on the LLaVA \cite{liu2023llava} and Shikra \cite{chen2023shikra} repository. Shraman and Rama were partially supported by a ONR MURI grant N00014-20-1-2787.

{
\small
\bibliographystyle{ieeenat_fullname}
\bibliography{main}
}


\newpage
\clearpage
\appendix
\counterwithin{figure}{section}
\numberwithin{table}{section}

\section{Radar Chart Figure 1 Details}

In this section, we explain the details of the radar chart in Figure 1, which summarizes the comparative performance of \model\ with MiniGPT-v2 \cite{chen2023minigptv2}, Ferret \cite{you2023ferret}, Shikra \cite{chen2023shikra} and GPT4RoI \cite{zhang2023gpt4roi}. None of these baselines address segmentation and multi-image tasks using a single framework. First, for illustrative purposes, we normalize each axis by the score achieved by \model, which turns the axes in the range $(0, 1]$. Next, we choose the origin of each axes suitably to distinctly separate the the inner and outer frames for better readability. For BoxQA, REC, and COCO Cap, the origin is at $0.97$, $0.96$, and $0.75$ normalized values, respectively. For all remaining axes, the origin is at $0.92$ normalized value. Finally, we annotate each vertex with absolute performance metric scores. The reported metric and split name for each axis are listed in Table \ref{tab:radar_metric}.

\begin{table}[!t]
\centering

\small
\setlength{\tabcolsep}{4pt}
\resizebox{0.7\columnwidth}{!}{\begin{tabular}{l | l}

\toprule 

\bf Axis & \bf Metric and Split \\

\midrule

\textcolor{red}{COCO Cap} & CIDEr on Karpathy test\\
\textcolor{red}{VQAv2} & Accuracy on val \\
\textcolor{red}{VCR} & Accuracy on val in Q $\to$ AR setup \\
\textcolor{red}{POPE} & F1 score on Random split \\
\textcolor{red}{HM} & Accuracy on test \\
\textcolor{red}{TextVQA} & Accuracy on test \\
\textcolor{blue}{REC} & Precision@IoU=0.5 on RefCOCO val \\
\textcolor{blue}{RES} & mIoU on RefCOCO val \\
\textcolor{blue}{GREC} & Precision on RefCOCO val \\
\textcolor{blue}{GRES} & gIoU on RefCOCO val \\
\textcolor{blue}{BoxQA} & Accuracy on Visual7W \\
\textcolor{OliveGreen}{NLVR2} & Accuracy on dev \\
\textcolor{Plum}{IconQA} & Accuracy on test \\
\textcolor{Plum}{iCoSeg} & Average Jaccard index ($\mathcal{J}$) on test \\

\bottomrule
\end{tabular}}
\vspace{-2mm}
\caption{\textbf{Details of the reported metrics and split information in every axis of the radar plot in Figure 1.} \textcolor{red}{Red}: Single-image coarse-level tasks, \textcolor{blue}{Blue}: Single-image region-level tasks, \textcolor{OliveGreen}{OliveGreen}: Multi-image coarse-level tasks, and \textcolor{Plum}{Plum}: Multi-image region-level tasks. } 
\label{tab:radar_metric}
\vspace{-3mm}
\end{table}

\vspace{-2mm}
\section{Adaptive Sampling Algorithm}

\begin{algorithm}[!t]\small
\caption{Gradient-aware Adaptive Sampling}
\label{alg:algo_sampling}
\begin{algorithmic}
    \Require Mask contour $\mathcal{C}$ \\
    Number of dense points $M$ \\
    Final number of sampling points $N$ (N $\ll$ M) \\
    $[p_{\mathrm{1}},\dots,p_{\mathrm{M}}]$ $\gets$ \textit{Uniform-Sample}($\mathcal{C}$) \Comment{Contour Discretization}
    \For{$i \in \{1,\dots,M\}$}
    \State $\vec{l_{\mathrm{1}}}$ = \textit{Join}($p_{\mathrm{i}}, p_{\mathrm{i-1}}$)
    \State $\vec{l_{\mathrm{2}}}$ = \textit{Join}($p_{\mathrm{i-1}}, p_{\mathrm{i+1}}$)
    \State $\theta_{\mathrm{i}}$ = $\angle \vec{l_{\mathrm{1}}}\vec{l_{\mathrm{2}}}$ \Comment{Gradient Calculation}

    \EndFor

    \State \text{Final}$_\text{points}$ $\gets$ []
    \State indices $\gets$ argsort($\theta_{\mathrm{i}\in\{1,\dots,M\}}$)[M-N:] \Comment{Sorting} 
    \For{$j \in \text{indices}$}
    \State $p_{\mathrm{j}}$ $\gets$ \textit{Quantize}($p_{\mathrm{j}}$)
    \State AddItem(\text{Final}$_\text{points}$, $p_{\mathrm{j}}$) \Comment{Quantization}
    \EndFor
    \State \text{Final}$_\text{points}$ is the final list of sampled points.
\end{algorithmic}
\end{algorithm}

The algorithm of the proposed gradient-aware adaptive sampling technique is given in Algorithm \ref{alg:algo_sampling}. Section 3.2 of the main manuscript provides details of each step. 


\section{\model\ vs Existing Region-level MLLMs}

\begin{table*}[!t]
\centering

\small
\setlength{\tabcolsep}{2pt}
\renewcommand{\arraystretch}{1.05}
\resizebox{0.65\textwidth}{!}{\begin{tabular}{@{} c l | c c c c c | c c c c @{}}

\toprule
& \multirow{3}{*}{\textbf{Model}} & \multicolumn{5}{c|}{\multirow{1}{*}{\textbf{Input Type}}} & \multicolumn{4}{c}{\multirow{1}{*}{\textbf{Output Type}}}  \\ 
\cmidrule{3-11}
& & \multirow{2}{1.15cm}{\centering Multiple Images} & \multirow{2}{*}{\centering Text} & \multirow{2}{*}{Points} & \multirow{2}{*}{Boxes} & \multirow{2}{*}{Masks} & \multirow{2}{*}{\centering Text} & \multirow{2}{*}{Points} & \multirow{2}{*}{Boxes} & \multirow{2}{*}{Masks} \\

& & & & & & & & & & \\ 

\midrule

\multirow{4}{1.2cm}{\centering Two-Stage} & Visual ChatGPT \cite{wu2023visual} & \textcolor{red}{\textcolor{red}{\ding{55}}} & \textcolor{RoyalBlue}{\ding{51}} & \textcolor{red}{\ding{55}} & \textcolor{red}{\ding{55}} & \textcolor{red}{\ding{55}} & \textcolor{RoyalBlue}{\ding{51}} & \textcolor{red}{\ding{55}} & \textcolor{RoyalBlue}{\ding{51}} & \textcolor{RoyalBlue}{\ding{51}}  \\

& BuboGPT \cite{zhao2023bubogpt}  & \textcolor{red}{\ding{55}} & \textcolor{RoyalBlue}{\ding{51}} & \textcolor{red}{\ding{55}} & \textcolor{red}{\ding{55}} & \textcolor{red}{\ding{55}} & \textcolor{RoyalBlue}{\ding{51}} & \textcolor{red}{\ding{55}} & \textcolor{RoyalBlue}{\ding{51}} & \textcolor{red}{\ding{55}} \\

& DetGPT \cite{pi2023detgpt} & \textcolor{red}{\ding{55}} & \textcolor{RoyalBlue}{\ding{51}} & \textcolor{red}{\ding{55}} & \textcolor{red}{\ding{55}} & \textcolor{red}{\ding{55}} & \textcolor{RoyalBlue}{\ding{51}} & \textcolor{red}{\ding{55}} & \textcolor{RoyalBlue}{\ding{51}} & \textcolor{red}{\ding{55}}\\

& LISA \cite{lai2023lisa} & \textcolor{red}{\ding{55}} & \textcolor{ProcessBlue}{\textcolor{RoyalBlue}{\ding{51}}} & \textcolor{red}{\ding{55}} & \textcolor{red}{\ding{55}} & \textcolor{red}{\ding{55}}  & \textcolor{RoyalBlue}{\ding{51}} & \textcolor{red}{\ding{55}} & \textcolor{red}{\ding{55}} & \textcolor{RoyalBlue}{\ding{51}}\\

\midrule

\multirow{12}{1.2cm}{\centering End-to-End} & LLaVa \cite{liu2023llava} & \textcolor{red}{\ding{55}} & \textcolor{RoyalBlue}{\ding{51}} & \textcolor{red}{\ding{55}} & \textcolor{red}{\ding{55}} & \textcolor{red}{\ding{55}} & \textcolor{RoyalBlue}{\ding{51}} & \textcolor{red}{\ding{55}} & \textcolor{red}{\ding{55}} & \textcolor{red}{\ding{55}} \\

& InstructBLIP \cite{instructblip} & \textcolor{red}{\ding{55}} & \textcolor{RoyalBlue}{\ding{51}} & \textcolor{red}{\ding{55}} & \textcolor{red}{\ding{55}} & \textcolor{red}{\ding{55}} & \textcolor{RoyalBlue}{\ding{51}} & \textcolor{red}{\ding{55}} & \textcolor{red}{\ding{55}} & \textcolor{red}{\ding{55}} \\

& GPT4RoI \cite{zhang2023gpt4roi} & \textcolor{red}{\ding{55}} & \textcolor{RoyalBlue}{\ding{51}} & \textcolor{red}{\ding{55}} & \textcolor{RoyalBlue}{\ding{51}} & \textcolor{red}{\ding{55}} & \textcolor{RoyalBlue}{\ding{51}} & \textcolor{red}{\ding{55}} & \textcolor{red}{\ding{55}} & \textcolor{red}{\ding{55}} \\

& KOSMOS-2 \cite{peng2023kosmos2} & \textcolor{red}{\ding{55}} & \textcolor{RoyalBlue}{\ding{51}} & \textcolor{red}{\ding{55}} & \textcolor{RoyalBlue}{\ding{51}} & \textcolor{red}{\ding{55}} & \textcolor{RoyalBlue}{\ding{51}} & \textcolor{red}{\ding{55}} & \textcolor{RoyalBlue}{\ding{51}} & \textcolor{red}{\ding{55}} \\

& VisionLLM \cite{wang2023visionllm} & \textcolor{red}{\ding{55}} & \textcolor{RoyalBlue}{\ding{51}} & \textcolor{red}{\ding{55}} & \textcolor{red}{\ding{55}} & \textcolor{red}{\ding{55}} & \textcolor{RoyalBlue}{\ding{51}} & \textcolor{red}{\ding{55}} & \textcolor{RoyalBlue}{\ding{51}} & \textcolor{RoyalBlue}{\ding{51}} \\

& Shikra \cite{chen2023shikra} & \textcolor{red}{\ding{55}} & \textcolor{RoyalBlue}{\ding{51}} & \textcolor{RoyalBlue}{\ding{51}} & \textcolor{RoyalBlue}{\ding{51}} & \textcolor{red}{\ding{55}} & \textcolor{RoyalBlue}{\ding{51}} & \textcolor{RoyalBlue}{\ding{51}} & \textcolor{RoyalBlue}{\ding{51}} & \textcolor{red}{\ding{55}} \\

& PVIT \cite{chen2023position} & \textcolor{red}{\ding{55}} & \textcolor{RoyalBlue}{\ding{51}} & \textcolor{red}{\ding{55}} & \textcolor{RoyalBlue}{\ding{51}} & \textcolor{red}{\ding{55}} & \textcolor{RoyalBlue}{\ding{51}} & \textcolor{red}{\ding{55}} & \textcolor{red}{\ding{55}} & \textcolor{red}{\ding{55}} \\

& CogVLM \cite{wang2023cogvlm} & \textcolor{red}{\ding{55}} & \textcolor{RoyalBlue}{\ding{51}} & \textcolor{red}{\ding{55}} & \textcolor{RoyalBlue}{\ding{51}} & \textcolor{red}{\ding{55}} & \textcolor{RoyalBlue}{\ding{51}} & \textcolor{red}{\ding{55}} & \textcolor{RoyalBlue}{\ding{51}} & \textcolor{red}{\ding{55}}\\

& COMM \cite{jiang2023comm} & \textcolor{red}{\ding{55}} & \textcolor{RoyalBlue}{\ding{51}} & \textcolor{red}{\ding{55}} & \textcolor{RoyalBlue}{\ding{51}} & \textcolor{red}{\ding{55}} & \textcolor{RoyalBlue}{\ding{51}} & \textcolor{red}{\ding{55}} & \textcolor{RoyalBlue}{\ding{51}} & \textcolor{red}{\ding{55}}\\

& MiniGPT-v2 \cite{chen2023minigptv2} & \textcolor{red}{\ding{55}} & \textcolor{RoyalBlue}{\ding{51}} & \textcolor{RoyalBlue}{\ding{51}} & \textcolor{RoyalBlue}{\ding{51}} & \textcolor{red}{\ding{55}} & \textcolor{RoyalBlue}{\ding{51}} & \textcolor{RoyalBlue}{\ding{51}} & \textcolor{RoyalBlue}{\ding{51}} & \textcolor{red}{\ding{55}}  \\

& Ferret \cite{you2023ferret} & \textcolor{red}{\ding{55}} & \textcolor{RoyalBlue}{\ding{51}} & \textcolor{RoyalBlue}{\ding{51}} & \textcolor{RoyalBlue}{\ding{51}} & \textcolor{RoyalBlue}{\ding{51}} & \textcolor{RoyalBlue}{\ding{51}} & \textcolor{RoyalBlue}{\ding{51}} & \textcolor{RoyalBlue}{\ding{51}} & \textcolor{red}{\ding{55}} \\

& \CC{}\model\ & \CC{}\textcolor{RoyalBlue}{\ding{51}} & \CC{}\textcolor{RoyalBlue}{\ding{51}} & \CC{}\textcolor{RoyalBlue}{\ding{51}} & \CC{}\textcolor{RoyalBlue}{\ding{51}} & \CC{}\textcolor{RoyalBlue}{\ding{51}} & \CC{}\textcolor{RoyalBlue}{\ding{51}} & \CC{}\textcolor{RoyalBlue}{\ding{51}} & \CC{}\textcolor{RoyalBlue}{\ding{51}} & \CC{}\textcolor{RoyalBlue}{\ding{51}} \\

\bottomrule
\end{tabular}}
\vspace{-2mm}
\caption{\textbf{Comparison of \model\ vs. existing general-purpose vision systems regarding input and output types.} \model\ supports all possible formats, including multiple images, natural language, points, bounding boxes, segmentation masks as inputs, and free-flowing text, points, boxes, and masks as output.}
\label{tab:comparison_baselines_inputs_outputs}
\vspace{-2mm}
\end{table*}

\begin{table*}[!t]
\centering

\small
\setlength{\tabcolsep}{2pt}
\renewcommand{\arraystretch}{1.08}
\resizebox{0.99\textwidth}{!}{\begin{tabular}{@{} c l | c c c | c c c c c c @{}}

\toprule
& \multirow{3}{*}{\textbf{Model}} & \multicolumn{3}{c|}{\multirow{1}{*}{\textbf{Image-level Tasks}}} & \multicolumn{6}{c}{\multirow{1}{*}{\textbf{Region-level Tasks}}}  \\ 
\cmidrule{3-11}
& & \multicolumn{2}{c}{\multirow{1}{*}{\centering \bf Single-image}} & \multicolumn{1}{c|}{\multirow{1}{*}{\centering \bf Multi-image}} & \multicolumn{5}{c}{\multirow{1}{*}{\centering \bf Single-image}} & \multicolumn{1}{c}{\multirow{1}{*}{\centering \bf Multi-image}} \\

& & \multirow{2}{1.5cm}{\centering VQAv2 \& Captioning} & \multirow{2}{*}{Reasoning} & \multirow{2}{*}{Reasoning} & \multirow{2}{*}{BoxQA} & \multirow{2}{*}{PointQA} & \multirow{2}{*}{Detection} & \multirow{2}{*}{Segmentation} & \multirow{2}{2cm}{\centering Multi-instance Segmentation} & \multirow{2}{*}{CoSeg} \\
& & & & & & & & & & \\

\midrule

\multirow{4}{1.2cm}{\centering Two-Stage} & Visual ChatGPT \cite{wu2023visual} & \textcolor{RoyalBlue}{\ding{51}} & \textcolor{RoyalBlue}{\ding{51}} & \textcolor{red}{\ding{55}} & \textcolor{red}{\ding{55}} & \textcolor{red}{\ding{55}} & \textcolor{RoyalBlue}{\ding{51}} & \textcolor{RoyalBlue}{\ding{51}} & \textcolor{RoyalBlue}{\ding{51}} & \textcolor{red}{\ding{55}}  \\

& BuboGPT \cite{zhao2023bubogpt} & \textcolor{RoyalBlue}{\ding{51}} & \textcolor{RoyalBlue}{\ding{51}} & \textcolor{red}{\ding{55}} & \textcolor{red}{\ding{55}} & \textcolor{red}{\ding{55}} & \textcolor{RoyalBlue}{\ding{51}} & \textcolor{red}{\ding{55}} & \textcolor{red}{\ding{55}} & \textcolor{red}{\ding{55}} \\

& DetGPT \cite{pi2023detgpt} & \textcolor{RoyalBlue}{\ding{51}} & \textcolor{RoyalBlue}{\ding{51}} & \textcolor{red}{\ding{55}} & \textcolor{red}{\ding{55}} & \textcolor{red}{\ding{55}} & \textcolor{RoyalBlue}{\ding{51}} & \textcolor{red}{\ding{55}} & \textcolor{red}{\ding{55}} & \textcolor{red}{\ding{55}}\\

& LISA \cite{lai2023lisa} & \textcolor{RoyalBlue}{\ding{51}} & \textcolor{RoyalBlue}{\ding{51}} & \textcolor{red}{\ding{55}} & \textcolor{red}{\ding{55}} & \textcolor{red}{\ding{55}} & \textcolor{red}{\ding{55}} & \textcolor{RoyalBlue}{\ding{51}} & \textcolor{red}{\ding{55}} & \textcolor{red}{\ding{55}} \\

\midrule

\multirow{12}{1.2cm}{\centering End-to-End} & LLaVa \cite{liu2023llava} & \textcolor{RoyalBlue}{\ding{51}} & \textcolor{RoyalBlue}{\ding{51}} & \textcolor{red}{\ding{55}} & \textcolor{red}{\ding{55}} & \textcolor{red}{\ding{55}} & \textcolor{red}{\ding{55}} & \textcolor{red}{\ding{55}} & \textcolor{red}{\ding{55}} & \textcolor{red}{\ding{55}} \\

& InstructBLIP \cite{instructblip} & \textcolor{RoyalBlue}{\ding{51}} & \textcolor{RoyalBlue}{\ding{51}} & \textcolor{red}{\ding{55}} & \textcolor{red}{\ding{55}} & \textcolor{red}{\ding{55}} & \textcolor{red}{\ding{55}} & \textcolor{red}{\ding{55}} & \textcolor{red}{\ding{55}} & \textcolor{red}{\ding{55}} \\

& GPT4RoI \cite{zhang2023gpt4roi} & \textcolor{RoyalBlue}{\ding{51}} & \textcolor{RoyalBlue}{\ding{51}} & \textcolor{red}{\ding{55}} & \textcolor{RoyalBlue}{\ding{51}} & \textcolor{red}{\ding{55}} & \textcolor{red}{\ding{55}} & \textcolor{red}{\ding{55}} & \textcolor{red}{\ding{55}} & \textcolor{red}{\ding{55}} \\

& KOSMOS-2 \cite{peng2023kosmos2} & \textcolor{RoyalBlue}{\ding{51}} & \textcolor{RoyalBlue}{\ding{51}} & \textcolor{red}{\ding{55}} & \textcolor{RoyalBlue}{\ding{51}} & \textcolor{red}{\ding{55}} & \textcolor{RoyalBlue}{\ding{51}} & \textcolor{red}{\ding{55}} & \textcolor{red}{\ding{55}} & \textcolor{red}{\ding{55}} \\

& VisionLLM \cite{wang2023visionllm} & \textcolor{RoyalBlue}{\ding{51}} & \textcolor{RoyalBlue}{\ding{51}} & \textcolor{red}{\ding{55}} & \textcolor{red}{\ding{55}} & \textcolor{red}{\ding{55}} & \textcolor{RoyalBlue}{\ding{51}} & \textcolor{RoyalBlue}{\ding{51}} & \textcolor{RoyalBlue}{\ding{51}} & \textcolor{red}{\ding{55}} \\

& Shikra \cite{chen2023shikra} & \textcolor{RoyalBlue}{\ding{51}} & \textcolor{RoyalBlue}{\ding{51}} & \textcolor{red}{\ding{55}} & \textcolor{RoyalBlue}{\ding{51}} & \textcolor{RoyalBlue}{\ding{51}} & \textcolor{RoyalBlue}{\ding{51}} & \textcolor{red}{\ding{55}} & \textcolor{red}{\ding{55}} & \textcolor{red}{\ding{55}}  \\

& PVIT \cite{chen2023position} & \textcolor{RoyalBlue}{\ding{51}} & \textcolor{RoyalBlue}{\ding{51}} & \textcolor{red}{\ding{55}} & \textcolor{RoyalBlue}{\ding{51}} & \textcolor{red}{\ding{55}} & \textcolor{red}{\ding{55}} & \textcolor{red}{\ding{55}} & \textcolor{red}{\ding{55}} & \textcolor{red}{\ding{55}} \\

& CogVLM \cite{wang2023cogvlm} & \textcolor{RoyalBlue}{\ding{51}} & \textcolor{RoyalBlue}{\ding{51}} & \textcolor{red}{\ding{55}} & \textcolor{RoyalBlue}{\ding{51}} & \textcolor{red}{\ding{55}} & \textcolor{RoyalBlue}{\ding{51}} & \textcolor{red}{\ding{55}} & \textcolor{red}{\ding{55}} & \textcolor{red}{\ding{55}} \\

& COMM \cite{jiang2023comm} & \textcolor{RoyalBlue}{\ding{51}} & \textcolor{RoyalBlue}{\ding{51}} & \textcolor{red}{\ding{55}} & \textcolor{RoyalBlue}{\ding{51}} & \textcolor{red}{\ding{55}} & \textcolor{RoyalBlue}{\ding{51}} & \textcolor{red}{\ding{55}} & \textcolor{red}{\ding{55}} & \textcolor{red}{\ding{55}} \\

& MiniGPT-v2 \cite{chen2023minigptv2} & \textcolor{RoyalBlue}{\ding{51}} & \textcolor{RoyalBlue}{\ding{51}} & \textcolor{red}{\ding{55}} & \textcolor{RoyalBlue}{\ding{51}} & \textcolor{RoyalBlue}{\ding{51}} & \textcolor{RoyalBlue}{\ding{51}} & \textcolor{red}{\ding{55}} & \textcolor{red}{\ding{55}} & \textcolor{red}{\ding{55}} \\

& Ferret \cite{you2023ferret} & \textcolor{RoyalBlue}{\ding{51}} & \textcolor{RoyalBlue}{\ding{51}} & \textcolor{red}{\ding{55}} & \textcolor{RoyalBlue}{\ding{51}} & \textcolor{RoyalBlue}{\ding{51}} & \textcolor{RoyalBlue}{\ding{51}} & \textcolor{red}{\ding{55}} & \textcolor{red}{\ding{55}} & \textcolor{red}{\ding{55}} \\

& \CC{}\model\ & \CC{}\textcolor{RoyalBlue}{\ding{51}} & \CC{}\textcolor{RoyalBlue}{\ding{51}} & \CC{}\textcolor{RoyalBlue}{\ding{51}} & \CC{}\textcolor{RoyalBlue}{\ding{51}} & \CC{}\textcolor{RoyalBlue}{\ding{51}} & \CC{}\textcolor{RoyalBlue}{\ding{51}} & \CC{}\textcolor{RoyalBlue}{\ding{51}} & \CC{}\textcolor{RoyalBlue}{\ding{51}} & \CC{}\textcolor{RoyalBlue}{\ding{51}} \\

\bottomrule
\end{tabular}}
\vspace{-2mm}
\caption{\textbf{Comparison of \model\ vs. existing general-purpose vision systems regarding supported tasks.} \model\ integrates a wide range of image-level and region-level vision-language reasoning and grounding tasks over single and multiple input images into a unified framework.}
\label{tab:comparison_baselines_tasks}
\vspace{-3.65mm}
\end{table*}

With the fast progress of region-level general-purpose vision systems, works such as GPT4RoI \cite{zhang2023gpt4roi}, Shikra \cite{chen2023shikra}, VisionLLM \cite{wang2023visionllm}, KOSMOS-2 \cite{peng2023kosmos2} and Ferret \cite{you2023ferret} resemble \model, as they also aim to unify tasks with different granularity in a unified system. Additional related works in this category includes PVIT \cite{chen2023position}, COMM \cite{jiang2023comm}, CogVLM \cite{wang2023cogvlm} and MiniGPT-v2 \cite{chen2023minigptv2}. Moreover, methods like Visual ChatGPT \cite{wu2023visual}, BuboGPT \cite{zhao2023bubogpt}, DetGPT \cite{pi2023detgpt}, and LISA \cite{lai2023lisa} employ external additional detection and segmentation modules to unify fine-grained tasks in a two-stage approach. Nevertheless, there exist clear differences between \model\ from existing methods. First, we present the first general-purpose system to support all possible input and output formats, e.g., multiple images, natural language, coordinate points, bounding boxes, segmentation masks as inputs, and free-flowing text, points, boxes, and masks as output. Table \ref{tab:comparison_baselines_inputs_outputs} shows a side-by-side comparison of input-output formats of all existing baselines. While Ferret supports boxes, points, and masks in the input, it can not generate a mask as output and, hence, can not address the segmentation task. On the other hand, VisionLLM can solve segmentation but cannot process points, boxes, and masks in input and can not solve REG, BoxQA, and PointQA. Second, unlike all existing works, \model\ supports multi-image input, enabling us to reason and ground over more than one image and solve tasks like NLVR and CoSeg. Our proposed instruction-guided image tokenizer module refines and compresses the global image embeddings of multiple images, helping \model\ to filter the necessary visual information required for the current task. Table \ref{tab:comparison_baselines_tasks} systematically illustrates the capability of \model\ to solve a wide range of image-level and region-level tasks over single and multiple input images compared to previous systems. Third, to efficiently convert segmentation masks into sequences, we propose a gradient-aware adaptive contour sampling scheme, which improves over previously used uniform sampling approach \cite{chen2021pix2seq, chen2022unified, zhu2022seqtr, liu2023polyformer} by $3-4$ mIoU scores on different segmentation benchmarks. Lastly, we collect a new training benchmark \data, containing $6.8$M training samples and propose a new task, AttCoSeg (\textbf{Att}ribute-level \textbf{Co}-\textbf{Seg}mentation) which addresses the lack of publicly-available multi-image region-level datasets. Our proposed system achieves stronger performance across $15$ different evaluation benchmarks, including mitigating object hallucination to a significant extent. 

\vspace{-1mm}
\section{Dataset Details} \label{sec:supp_dataset_details}
This section provides additional details of our training and evaluation datasets.

\vspace{1mm}
\noindent \textbf{COCO Captioning:} Captions for the COCO dataset \cite{lin2014microsoft} were sourced from Amazon's Mechanical Turk (AMT), with workers adhering to specified guidelines to ensure caption quality. The dataset includes 330,000 images, divided into training, validation, and test categories. These categories comprise 413,915 captions for 82,783 images in training, 202,520 captions for 40,504 images in validation, and 379,249 captions for 40,775 images in the test set.

\vspace{1mm}
\noindent \textbf{VQAv2:} VQAv2 dataset \cite{antol2015vqa} contains a collection of over 200,000 images, each paired with a portion of the more than 1.1 million questions asked, gathering in total over 11 million responses. The questions cover a wide range, from simple yes/no and counting queries to more complex open-ended ones.

\vspace{1mm}
\noindent \textbf{RefCOCO \& RefCOCO+:} The RefCOCO and RefCOCO+ datasets \cite{liu2017referring} were created through a two-player game mechanism \cite{yu2016modeling}. RefCOCO features 142,209 descriptive expressions for 50,000 objects across 19,994 images, whereas RefCOCO+ includes 141,564 expressions for 49,856 objects in 19,992 images. Both datasets are divided into training, validation, and two test sets – Test A and Test B. Test A focuses on images with multiple people. At the same time, Test B features images with multiple instances of all other objects. A key difference between the two datasets is that RefCOCO+ excludes location words from its expressions, making it more complex than RefCOCO. We perform referring expression comprehension (REC) and referring expression segmentation (RES) tasks on the RefCOCO and RefCOCO+ datasets.

\vspace{1mm}
\noindent \textbf{RefCOCOg:} The RefCOCOg dataset was assembled using Amazon Mechanical Turk, where participants were tasked with crafting natural language descriptions for objects. It comprises 85,474 expressions for 54,822 objects in 26,711 images. Notably, the expressions in RefCOCOg are longer and more intricate, averaging 8.4 words, in contrast to the more concise expressions in RefCOCO and RefCOCO+, which average 3.5 words. This complexity makes RefCOCOg a more challenging dataset. We utilize the UMD partition \cite{nagaraja2016modeling} of RefCOCOg, as it provides both validation and testing sets, and there is no overlap between training and validation images. We address both REC and RES tasks on RefCOCOg.

\vspace{1mm}
\noindent \textbf{gRefCOCO:} The gRefCOCO dataset \cite{liu2023gres, he2023grec} empowers generalized referring expression comprehension (GREC) and generalized referring expression segmentation (GRES) tasks, which address the limitations of classical REC and RES problem where there is always one target object. In contrast, GREC and GRES allow expressions to refer to an arbitrary number of target objects, including multi-target and no-target scenarios, and help bring referring segmentation into more realistic scenarios with advanced usages. The gRefCOCO dataset contains 278,232 expressions, including 80,022 multi-target and 32,202 no-target expressions, referring to 60,287 distinct instances in 19,994 images. Masks and bounding boxes for all target instances are given. Some of the single-target expressions of gRofCOCO are inherited from RefCOCO. We perform both GREC and GRES using the gRefCOCO dataset.

\vspace{1mm}
\noindent \textbf{Flickr:} The Flickr30K Entities dataset \cite{plummer2015flickr30k} is a pioneering collection in the field of grounded captioning. It includes 31,783 images paired with 158,000 caption annotations. Each caption is carefully annotated, linking every noun phrase to a manually outlined referential bounding box. The dataset features a total of 276,000 such annotated bounding boxes, offering a rich resource for image and language processing research. We use Flickr dataset during training for spot captioning task, where we instruct the model to generate a caption of the input image, and locate all the objects in the images by drawing bounding boxes. 

\vspace{1mm}
\noindent \textbf{Visual Genome:} The Visual Genome dataset \cite{krishna2017visual} is a key resource for understanding the complex relationships within images. It contains over 100,000 images, with each image extensively annotated to capture an average of 21 objects, 18 attributes, and 18 inter-object relationships. A distinctive feature of this dataset is the alignment of objects, attributes, relationships, and region descriptions with the standardized WordNet terminologies. This alignment makes it particularly useful for tasks like Region Description and Entity Recognition. Each annotated region in the dataset is accompanied by descriptive text, providing a wealth of data for image understanding and semantic modeling. For referring expression generation (REG) purposes, we utilize a subset of this dataset, which includes around 180,138 region-caption pairs.

\vspace{1mm}
\noindent \textbf{VCR:} The Visual Commonsense Reasoning (VCR) dataset \cite{zellers2019recognition} contains 290,000 multiple-choice questions derived from 110,000 movie scenes. Each scene is paired with a question demanding common-sense reasoning, an answer, and a rationale for that answer. The unique aspect of VCR is its requirement for models to not only provide answers to complex visual questions but also to explain their reasoning. This dataset encompasses two sub-tasks: Question Answering (Q $\to$ A), where the model selects the correct answer from four options, and answer justification (QA $\to$ R), where the model, given a question and its correct answer, must choose the most fitting rationale from four options. Model performance in VCR is assessed using the Q $\to$ AR metric, which measures the accuracy of both answering questions and providing the correct justifications.

\vspace{1mm}
\noindent \textbf{LLaVa:} The LLaVA-Instruct-150K\footnote{\url{https://huggingface.co/datasets/liuhaotian/LLaVA-Instruct-150K}} \cite{liu2023llava} is a collection of 158K unique language-image instruction-following samples in total, including 58K in conversations, 23K in the detailed description, and 77k in complex reasoning, respectively. We incorporate the LLaVa dataset during the training of our model. 

\vspace{1mm}
\noindent \textbf{LookTwiceQA:} The LookTwiceQA \cite{mani2020point} dataset contains two different tasks - PointQA and BoxQA. The questions are in three different templates - ($i$) What color is this [region]? ($ii$) What shape is this [region]? and ($iii$) What action is this [region] doing? The question contains either an input point or a box with three different granularity of objects - any object, superclass, and object class. The train set contains 40,409 questions across 12,867 images, and the test-dev set contains 5,673 questions across 1,838 images.

\vspace{1mm}
\noindent \textbf{Visual7W:} The Visual7W dataset \cite{zhu2016visual7w} is primarily tailored for Visual Question Answering (VQA) tasks, featuring a specialized dataset for region-level QA. In Visual7W, models encounter an image paired with a "which"-type question, for instance, "Which one is the orange in the fruit basket?". Participants are provided with four bounding boxes in the image and must choose the correct one as the answer. The Visual7W dataset comprises 25,733 images and 188,068 such questions.

\vspace{1mm}
\noindent \textbf{TextVQA:} TextVQA \cite{singh2019towards} is a QA dataset containing 45,336 questions based on 28,408 images, designed to challenge models in detecting, interpreting, and reasoning about text present in images to generate accurate answers. We use the TestVQA dataset as an unseen evaluation benchmark. 

\vspace{1mm}
\noindent \textbf{IconQA:} IconQA \cite{lu2021iconqa} measures models' abstract diagram understanding and comprehensive cognitive reasoning abilities. We use the test set of its multi-text-choice task, containing 6,316 samples, as an unseen evaluation benchmark.

\vspace{1mm}
\noindent \textbf{Hateful Memes (HM):} The hateful memes dataset \cite{kiela2020hateful}, containing more than 10,000 image samples, is a binary classification dataset to justify whether a meme contains hateful content. The memes were selected in such a way that strictly unimodal classifiers would struggle to classify them correctly. We use the HM dataset as an unseen evaluation benchmark.

\vspace{1mm}
\noindent \textbf{POPE:} The POPE evaluation benchmark \cite{li2023evaluating} evaluates the sevearity of object hallucination problem in MLLMs. POPE consists of three different test splits - popular, random, and adversarial- containing around 3,000 samples. Given an image and a question, "Is there a $<$object$>$ in the image?" the model has to answer with 'yes' or 'no.' 

\vspace{1mm}
\noindent \textbf{NLVR2:} The Natural Language for Visual Reasoning (NLVR2) corpora, containing 107,292 samples, determine whether a sentence is true about a pair of input images. The data was collected through crowdsourcing, and solving the task requires reasoning about sets of objects, comparisons, and spatial relations. 

\vspace{1mm}
\noindent \textbf{CoSeg:} We use three datasets for object co-segmentation task - PASCAL VOC2010 \cite{faktor2013co}, MSRC \cite{winn2005object} and iCoSeg \cite{batra2010icoseg}. PASCAL contains a total of 1,037 images of 20 object classes. MSRC includes seven classes: bird, car, cat, cow, dog, plane, and sheep. Each class contains ten images. iCoseg dataset consists of 643 images from 38 categories. Large variances of viewpoints and deformations are present in this dataset.

\vspace{1mm}
\noindent \textbf{AttCoSeg:} Since the existing object co-segmentation datasets \cite{faktor2013co, winn2005object, batra2010icoseg} are small-scale and simple to solve, we construct a more challenging larger-scale multi-image region-level dataset. We use Group-wise RES \cite{wu2023advancing} annotations to sample high-quality images containing objects with similar fine-grained attributes (shape, color, size, position). We refer to such images as positives. While training \model, we input these positive image pairs and ask the model to segment the object with common traits in both of them. We name this task attribute-level co-segmentation (AttCoSeg), which contains over 804k training samples, and help \model\ to gain significant generalized reasoning and grounding ability over multiple input images.

\section{Examples Instructions for Different Tasks}

Section 5.1 discusses transforming public datasets like REC, RES, GREC, and GRES into instruction-following format by employing meticulously crafted task templates. These templates are detailed in Table \ref{tab:instructions}. We have included only 2-3 examples for each task for brevity. We manually write one example description of each task and resort to GPT-3.5 \cite{brown2020language} to create hundreds of variations. During training, we randomly pick one instruction for each sample.

\begin{table*}[!t]
\centering

\small
\setlength{\tabcolsep}{4pt}
\resizebox{\textwidth}{!}{\begin{tabular}{l | p{15cm}}

\toprule 

\bf Task & \bf Example Instructions  \\ 

\midrule

\multirow{3}{*}{Captioning} & $\bullet$ Can you give me a brief description of this image $<$image$>$? \\
& $\bullet$ Give me a short description of the picture $<$image$>$. \\
& $\bullet$ What's happening in the image $<$image$>$ at a glance? \\
\midrule
\multirow{3}{*}{VQAv2} & $\bullet$ Looking at the image $<$image$>$, can you quickly answer my question: $<$question$>$. \\
& $\bullet$ After examining the image $<$image$>$, can you provide a brief response to the following question: $<$question$>$. \\
& $\bullet$ Considering the image $<$image$>$, please provide a straightforward answer to $<$question$>$. \\
\midrule
\multirow{4}{*}{REC} & $\bullet$ Locate the object described by $<$expr$>$ in $<$image$>$. There's just one specific object. Provide the outcome using the [x$_0$, y$_0$, x$_1$, y$_1$] arrangement, showing the upper-left and lower-right box positions. \\
& $\bullet$ Find the location of the item referenced in $<$expr$>$ within $<$image$>$. We're referring to a single item. Output the result in [x$_0$, y$_0$, x$_1$, y$_1$] arrangement, showing the upper-left and lower-right bounding box corners. \\
\midrule
\multirow{4}{*}{RES} & $\bullet$ Tell me where $<$expr$>$ is located in $<$image$>$. There's only one object. Provide the coordinates of 32 points on the object's outline. Present the result in [x$_0$, y$_0$, x$_1$, y$_1$, ..., x$_{31}$, y$_{31}$] format. \\
& $\bullet$ What is $<$expr$>$'s location within $<$image$>$? There's just one thing to consider. Share the coordinates of 32 uniform points on the object's edge. Show it in [x$_0$, y$_0$, x$_1$, y$_1$, ..., x$_{31}$, y$_{31}$] format. \\
\midrule
\multirow{6}{*}{GREC} & $\bullet$ Recognize all objects indicated by $<$expr$>$ in $<$image$>$. If no object is located, return an empty string. If one or more objects are located, output the bounding boxes as [x$_0$, y$_0$, x$_1$, y$_1$], indicating the top-left and bottom-right corner points. Use $<$bsep$>$ to differentiate multiple bounding boxes. \\
& $\bullet$ Pinpoint all items referenced by $<$expr$>$ in $<$image$>$. If no object is detected, return an empty string. If one or more target objects are found, provide the bounding boxes as [x$_0$, y$_0$, x$_1$, y$_1$], signifying the top-left and bottom-right corner points. Use $<$bsep$>$ to separate multiple bounding boxes. \\
\midrule
\multirow{6}{*}{GRES} & $\bullet$ Find all items indicated by $<$expr$>$ within $<$image$>$. If no target object is recognized, produce an empty string. If one or more target objects are identified, output the coordinates of 32 points along each object's contour. Display each object mask in [x$_0$, y$_0$, x$_1$, y$_1$, ..., x$_{31}$, y$_{31}$] format. Use $<$msep$>$ to distinguish multiple objects. \\
& $\bullet$ Recognize all referenced items via $<$expr$>$ in $<$image$>$. If no target object is found, generate an empty string. If one or more target objects are found, present the coordinates of 32 points along each object's edge. Show each object mask in [x$_0$, y$_0$, x$_1$, y$_1$, ..., x$_{31}$, y$_{31}$] format. Utilize $<$msep$>$ to distinguish multiple objects. \\
\midrule
\multirow{3}{*}{REG} & $\bullet$ Please generate a unique description for the area $<$objs$>$ displayed in the image $<$image$>$.\\
& $\bullet$ What can you tell me about the area $<$objs$>$ in the image $<$image$>$ that sets it apart from the rest? \\
& $\bullet$ How does the area $<$objs$>$ in $<$image$>$ stand out uniquely from the rest? \\
\midrule
\multirow{6}{*}{NLVR} & $\bullet$ Between the left image $<$image$>$ and the right image $<$image$>$, could you tell me if the answer to $<$question$>$ is True or False?\\
& $\bullet$ Reviewing both the left image $<$image$>$ and the right image $<$image$>$, would you reckon $<$question$>$ is True or False? \\
& $\bullet$ Given the left image $<$image$>$ and the right image $<$image$>$, can you answer my query: $<$question$>$? Respond in True or False. \\
\midrule
\multirow{4}{1.5cm}{Spot Captioning} & $\bullet$ Please provide a holistic description of the image $<$image$>$ and output the position for each mentioned object in the format [x$_0$, y$_0$, x$_1$, y$_1$] representing top-right and bottom-left corners of the bounding box. \\
& $\bullet$ Present a thorough insight into $<$image$>$ and output every object's position using [x$_0$, y$_0$, x$_1$, y$_1$], representing the bounding box's top-right and bottom-left corners. \\
\midrule
\multirow{4}{*}{CoSeg} & $\bullet$ Find the common object in the input images $<$image$>$. There's only one common object. Display each object's mask in [x$_0$, y$_0$, x$_1$, y$_1$, ..., x$_{31}$, y$_{31}$] format. Utilize $<$msep$>$ to tell the masks apart. \\
& $\bullet$ Locate the common thing in the input images $<$image$>$. Only one common thing will be there. Present each thing's mask in [x$_0$, y$_0$, x$_1$, y$_1$, ..., x$_{31}$, y$_{31}$] style. Use $<$msep$>$ to differentiate the two masks.\\
\midrule
\multirow{6}{*}{AttCoSeg} & $\bullet$ Find the two images which have a common object with matching attributes (shape, color, size, position), and segment it in both images. Show object mask in [x$_0$, y$_0$, x$_1$, y$_1$, ..., x$_{31}$, y$_{31}$] style in both pictures. Make use of $<$msep$>$ to tell apart the two masks. \\
& $\bullet$ Which input images have a mutual item with common attributes (shape, color, size, position)? Segment it in both images. Display object mask using [x$_0$, y$_0$, x$_1$, y$_1$, ..., x$_{31}$, y$_{31}$] format in both images. Apply $<$msep$>$ to differentiate the two masks. \\

\bottomrule
\end{tabular}}
\vspace{-2mm}
\caption{\textbf{Examples of instructions} for different tasks used by \model\ to convert them into instruction-following format.} 
\label{tab:instructions}
\vspace{-5mm}
\end{table*}


\section{Additional Ablation Study} \label{sec:supp_additional_ablations}

In this section, we conduct additional ablation experiments on training dataset, and the image encoder. 

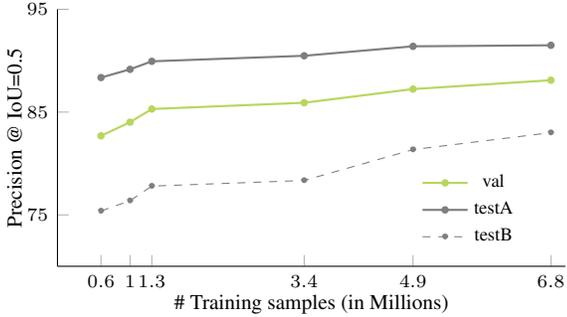
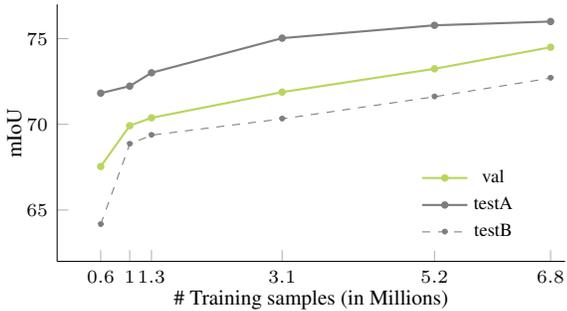
\begin{figure}[!t]
    \centering
    \hspace{-0em}
    \begin{subfigure}[b]{\columnwidth}
        \resizebox{!}{!}{
            \begin{tikzpicture}
	\begin{axis} [
        width=\textwidth,
        height=.6\textwidth,
		axis x line*=bottom,
		axis y line*=left,
		legend pos=north east,
		ymin=70, ymax=95,
		xmin=0, xmax=7,
		xticklabel={\pgfmathparse{\tick}\pgfmathprintnumber{\pgfmathresult}},
		xtick={0.6, 1., 1.3, 3.4, 4.9, 6.8},
		ytick={75,85,95},
        xlabel={\footnotesize{\# Training samples (in Millions)}},
        ylabel={\footnotesize{Precision @ IoU=0.5}},
        xlabel shift=-4pt,
        ylabel shift=-5pt,
		width=\columnwidth,
		legend style={cells={align=left}},
		label style={font=\normalsize},
		tick label style={font=\scriptsize},
		legend style={draw=none,at={(0.70,0.22)},anchor=west},
		]

        \addplot[mark=*,mark options={scale=0.5, fill=Dark},style={thick},Dark] plot coordinates {
			(0.6, 82.69)
            (1.0, 84.02)
			(1.3, 85.30)
			(3.4, 85.91)
            (4.9, 87.24)
            (6.8, 88.1)
		};
        \addlegendentry{\scriptsize{val}}

        \addplot[mark=*,mark options={scale=0.5, fill=Salmon},style={thick},Salmon] plot coordinates {
			(0.6, 88.36)
            (1.0, 89.16)
			(1.3, 89.94)
			(3.4, 90.48)
            (4.9, 91.40)
            (6.8, 91.5)
		};
        \addlegendentry{\scriptsize{testA}}

        \addplot[mark=*,mark options={scale=0.5, fill=Salmon},style={dashed},Salmon] plot coordinates {
			(0.6, 75.38)
            (1.0, 76.38)
			(1.3, 77.80)
			(3.4, 78.35)
            (4.9, 81.35)
            (6.8, 83.0)
		};
        \addlegendentry{\scriptsize{testB}}

	\end{axis}
\end{tikzpicture}%
        }
        \caption{\textbf{Performance of REC on RefCOCO with varying training samples.} We report the performance in terms of precision at IoU = 0.5, i.e., the prediction is deemed correct if its intersection over union (IoU) with the ground-truth box is larger than 0.5.}
        \vspace{2mm}
        \label{fig:ablation_dataset_REC}
    \end{subfigure}
 \hspace{0em}
    \begin{subfigure}[b]{\columnwidth}
        \resizebox{!}{!}{
            \begin{tikzpicture}
	\begin{axis} [
        width=\textwidth,
        height=.6\textwidth,
		axis x line*=bottom,
		axis y line*=left,
		legend pos=north east,
		ymin=62, ymax=77,
		xmin=0, xmax=7,
		xticklabel={\pgfmathparse{\tick}\pgfmathprintnumber{\pgfmathresult}},
		xtick={0.6, 1.0, 1.3, 3.1, 5.2, 6.8},
		ytick={65,70,75},
        xlabel={\footnotesize{\# Training samples (in Millions)}},
        ylabel={\footnotesize{mIoU}},
        xlabel shift=-4pt,
        ylabel shift=-5pt,
		width=\columnwidth,
		legend style={cells={align=left}},
		label style={font=\normalsize},
		tick label style={font=\scriptsize},
		legend style={draw=none,at={(0.70,0.22)},anchor=west},
		]

        \addplot[mark=*,mark options={scale=0.5, fill=Dark},style={thick},Dark] plot coordinates {
			(0.6, 67.54)
            (1.0, 69.92)
			(1.3, 70.38)
			(3.1, 71.88)
            (5.2, 73.24)
            (6.8, 74.5)
		};
        \addlegendentry{\scriptsize{val}}

        \addplot[mark=*,mark options={scale=0.5, fill=Salmon},style={thick},Salmon] plot coordinates {
			(0.6, 71.82)
            (1.0, 72.23)
			(1.3, 73.01)
			(3.1, 75.03)
            (5.2, 75.78)
            (6.8, 76.0)
		};
        \addlegendentry{\scriptsize{testA}}

        \addplot[mark=*,mark options={scale=0.5, fill=Salmon},style={dashed},Salmon] plot coordinates {
			(0.6, 64.16)
            (1.0, 68.85)
			(1.3, 69.37)
			(3.1, 70.32)
            (5.2, 71.61)
            (6.8, 72.7)
		};
        \addlegendentry{\scriptsize{testB}}

	\end{axis}
\end{tikzpicture}%
        }
        \caption{\textbf{Performance of RES on RefCOCO with varying number of training samples.} We report the performance in terms of mIoU score.}
        \label{fig:ablation_dataset_RES}
    \end{subfigure}
\vspace{-3mm}
\caption{\textbf{Ablation on the number of training samples on the REC and RES task performance.} We start with only RES and REC datasets and gradually append datasets from other tasks using proper instructions. Increasing the number of samples helps produce better performance, showing the usefulness of an end-to-end, cohesive, and unified system where different tasks help improve each other.}
\label{fig:ablation_dataset_REC_RES}
\vspace{-5mm}
\end{figure}

\vspace{1mm}
\noindent \textbf{Size of training dataset:} We study the effect of increasing training samples for REC and RES tasks in Figure \ref{fig:ablation_dataset_REC_RES}. We start with REC and REG training datasets for the REC task in Figure \ref{fig:ablation_dataset_REC}, resulting in $0.6$M training samples. We train \model\ for two epochs in stage 1, setting all hyperparameters unchanged. In this setup, we observe a REC val score of $82.7$\%. Next, we add Visual Genome data to the training corpus, which results in a total of $1$M samples, and re-train the model. Now, REC val accuracy increases to $84.0$\%. Similarly, appending PointQA data in the training corpus increases the performance by $1.3$\%, and appending LLaVa, Flickr, VQAv2, and COCO caption data yields a gain of another $0.7$\%. Finally, the $6.8$M training corpus produces a final REC val accuracy of $88.1$\%. Hence, we observe that datasets from other image-level and region-level tasks help improve the performance of the REC task, which is the benefit of unified end-to-end training. We also see similar observations for the RES in Figure \ref{fig:ablation_dataset_RES}. Such a phenomenon also proves the scalability of our approach, which is important for large-scale unified training.

\begin{table}[!t]
\vspace{2mm}
\centering

\small
\setlength{\tabcolsep}{4pt}
\resizebox{\columnwidth}{!}{\begin{tabular}{l | c c c c c c c}

\toprule

\multirow{2}{0.5cm}{\bf Method} & \multicolumn{1}{c}{\bf Cap.} & \multicolumn{3}{c}{\bf RES Ref} & \multicolumn{1}{c}{\bf VCR} & \multicolumn{1}{c}{\bf iCoSeg} & \multicolumn{1}{c}{\bf NLVR} \\ 

& CIDEr & val & testA & testB & Q $\to$ AR & Av. $\mathcal{J}$ & dev \\

\midrule

\rowcolor{Light}
\model-13B & \bf 128.4 & \bf 76.2 & \bf 77.7 & \bf 73.9 & \bf 79.1 & \bf 95.1 & \bf 80.8 \\
w/ CLIP-ViT-L/14 & 127.9 & 75.5 & 76.3 & 72.1 & 79.3 & 94.7 & 80.2 \\
w/ CLIP-ViT-L/14-336px & \bf 128.4 & 76.0 & \bf 77.7 & 73.6 & 79.3 & 95.1 & 80.5 \\
w/ CLIP-ViT-B/16 & 127.6 & 75.1 & 76.3 & 72.0 & 79.0 & 94.8 & 79.8 \\

\bottomrule
\end{tabular}}
\vspace{-2mm}
\caption{\textbf{Ablation with different image encoders.} By default, \model\ uses EVA-CLIP \cite{sun2023eva} pre-trained on LAION-$400$M \cite{schuhmann2021laion}. We observe a small performance drop when using other image encoders.} 
\label{tab:image_encoder_ablation}
\vspace{-4mm}
\end{table}

\vspace{1mm}
\noindent \textbf{Image encoder:} Next, we ablate different image encoders in Table \ref{tab:image_encoder_ablation}. We observe the best performance across most tasks with EVA\footnote{\url{https://huggingface.co/QuanSun/EVA-CLIP/blob/main/EVA01_CLIP_g_14_psz14_s11B.pt}} \cite{sun2023eva}, while the CLIP-ViT-L/14-336px\footnote{\url{https://huggingface.co/openai/clip-vit-large-patch14-336}} follows closely. We use EVA-CLIP in our final model because the QFormer \cite{li2023blip2} pre-trained in InstructBLIP \cite{instructblip} uses EVA-CLIP, and it results in best compatibility with the instruction-guided image tokenizer module in our system.

\section{Error Analysis}
\begin{figure*}[!t]
    \centering
    \includegraphics[width=\textwidth]{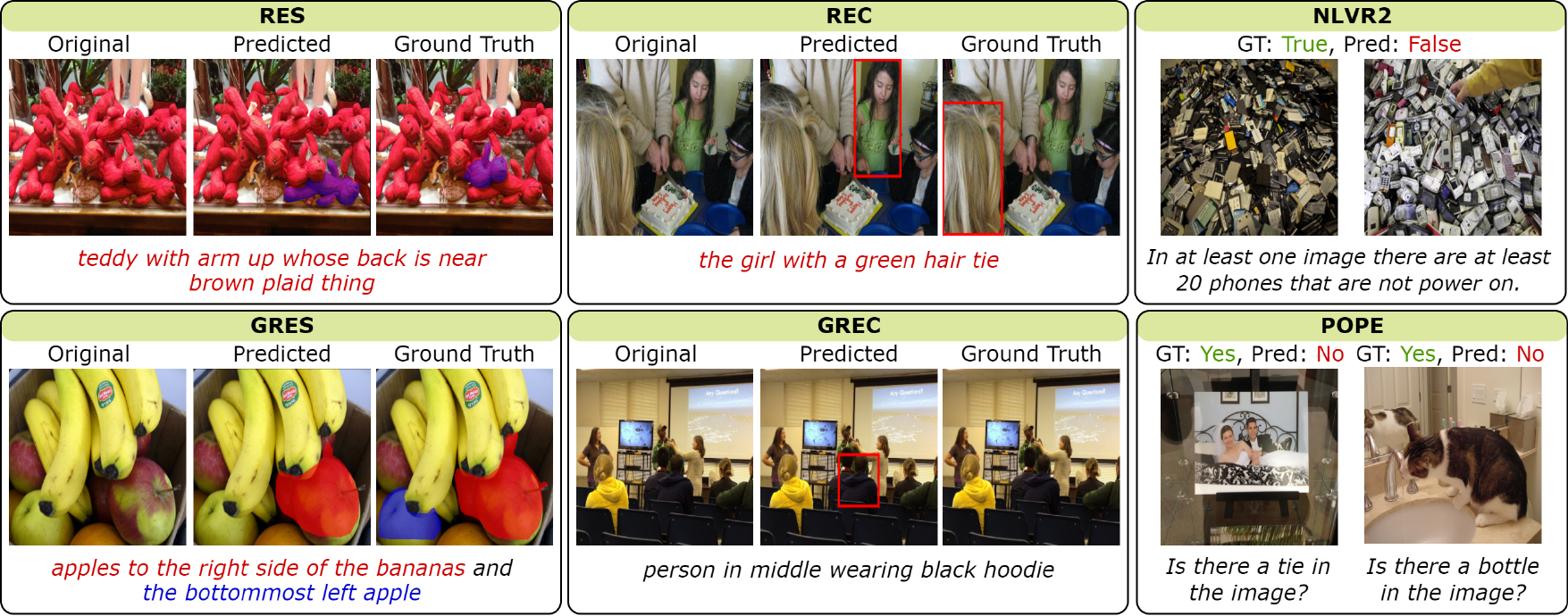}
    \vspace{-6mm}
    \caption{\textbf{Limitations of our method:} Tiny and obscured objects, especially in cluttered and low-light environments, are hard to be accurately grounded. \model\ fails in such tough samples, which are even difficult to comprehend by humans.}
    
    \label{fig:appendix_error_analysis}
\end{figure*}

Although \model\ learns impressive reasoning and grounding capability across many different benchmarks, there are still some cases where the model fails to identify small and obscured objects, especially in cluttered environments. Figure \ref{fig:appendix_error_analysis} shows seven such failure cases. In the RES example, the object ``\textit{teddy with arm up whose back in near brown plaid thing}'' is hard to comprehend even for humans, and thus, \model\ can not identify the correct ``\textit{teddy}'' the expression is referring to. In the REC example, the ``\textit{green hair tie}'' is tiny and only visible when zoomed into the picture. \model\ fails to identify the girl who is wearing it. In the GREC example, in low-light conditions, the blue hoodie appears to be black, and \model\ wrongly outputs a bounding box, whereas the ground truth is no matching object. Similarly, in the NLVR2, GRES, and POPE examples, \model\ fails to recognize hindered and cluttered objects. We believe that more robust image features will alleviate such failure cases in the future. Moreover, similar to many LLMs, \model\ has the potential to generate harmful and unsafe outputs, which is also an active research topic.   

\section{Additional Qualitative Results} \label{sec:supp_additional_qualitative_results}
\begin{figure*}[!t]
    \centering
    \includegraphics[width=\textwidth]{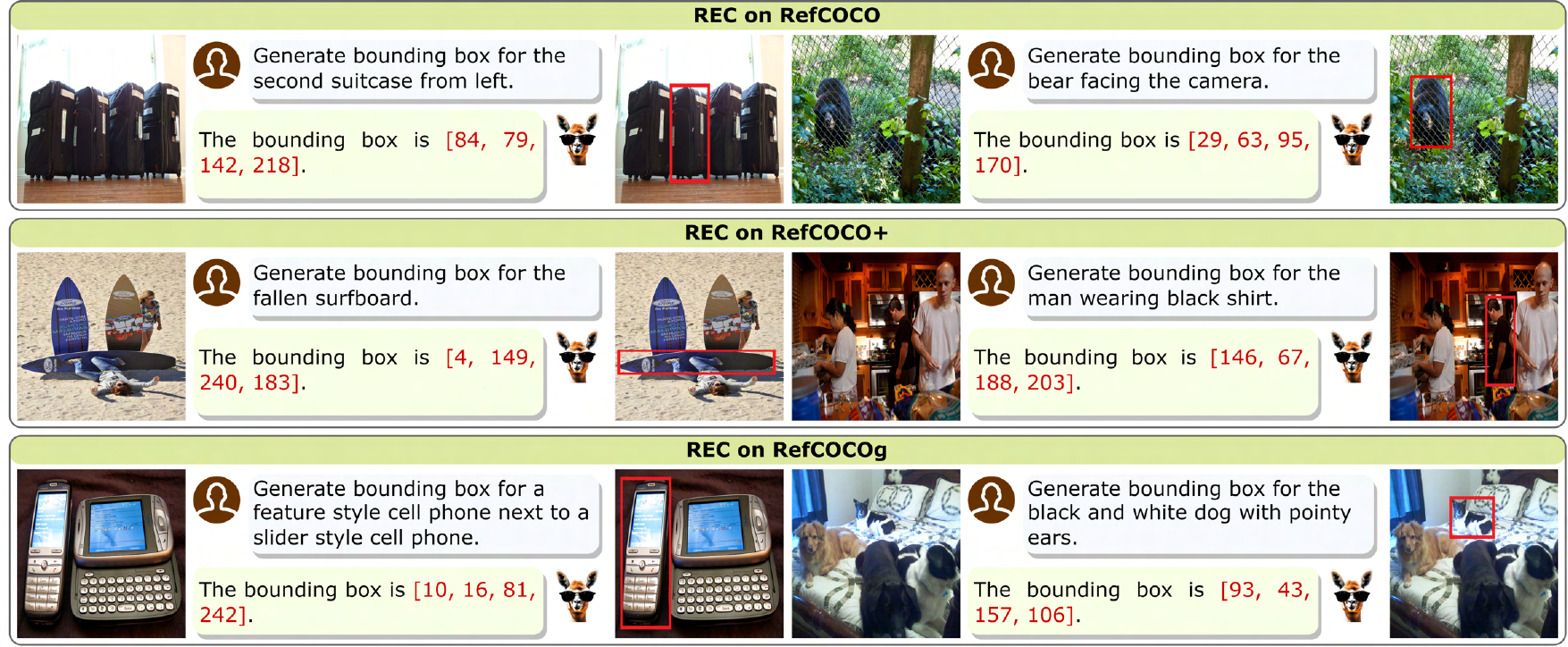}
    \vspace{-6mm}
    \caption{\textbf{Referring Expression Comprehension (REC) on RefCOCO, RefCOCO+ and RefCOCOg by \model-13B.} REC aims to generate a bounding box around a single object described by a referring expression.}
    \label{fig:appendix_rec}
\end{figure*}

\begin{figure*}[!t]
    \centering
    \includegraphics[width=\textwidth]{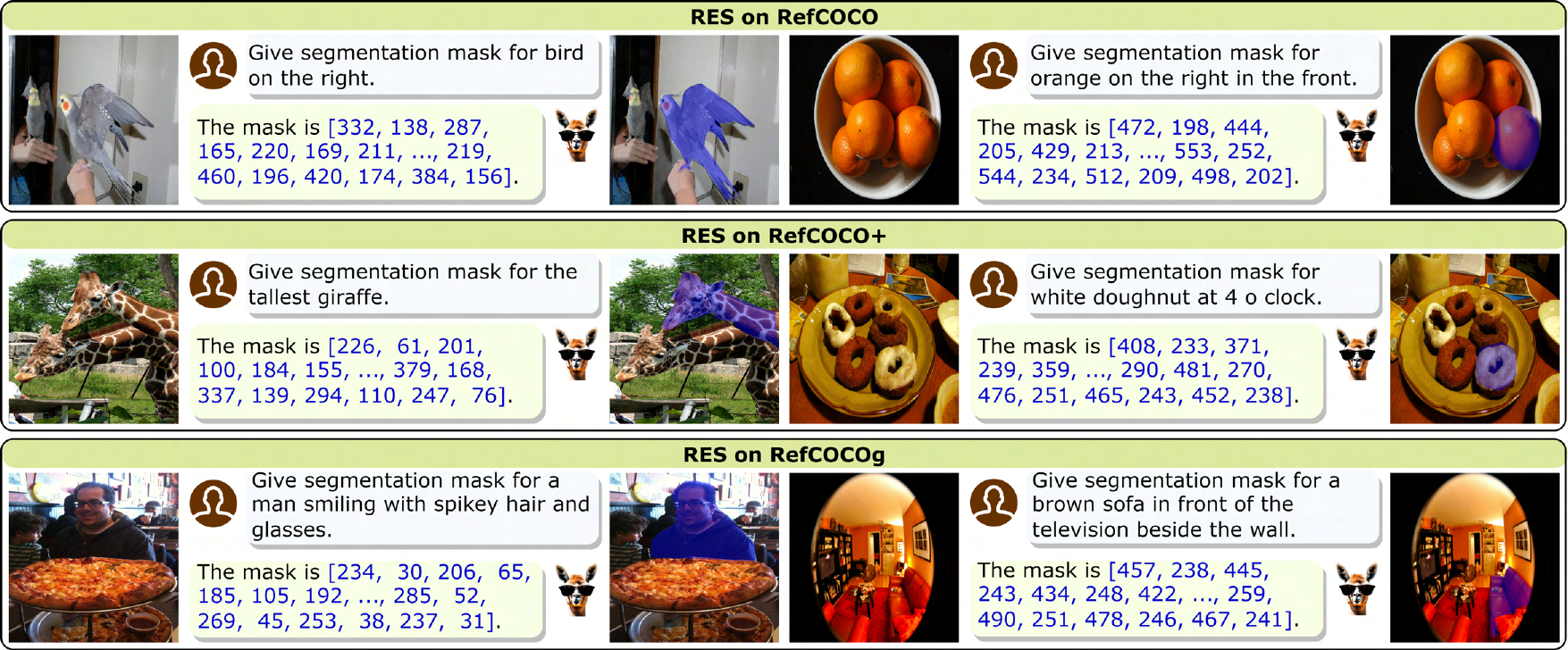}
    \vspace{-6mm}
    \caption{\textbf{Referring Expression Segmentation (RES) on RefCOCO, RefCOCO+ and RefCOCOg by \model-13B.} RES aims to segment a single object described by a referring expression.}
    \label{fig:appendix_res}
\end{figure*}

\begin{figure*}[!t]
    \centering
    \includegraphics[width=\textwidth]{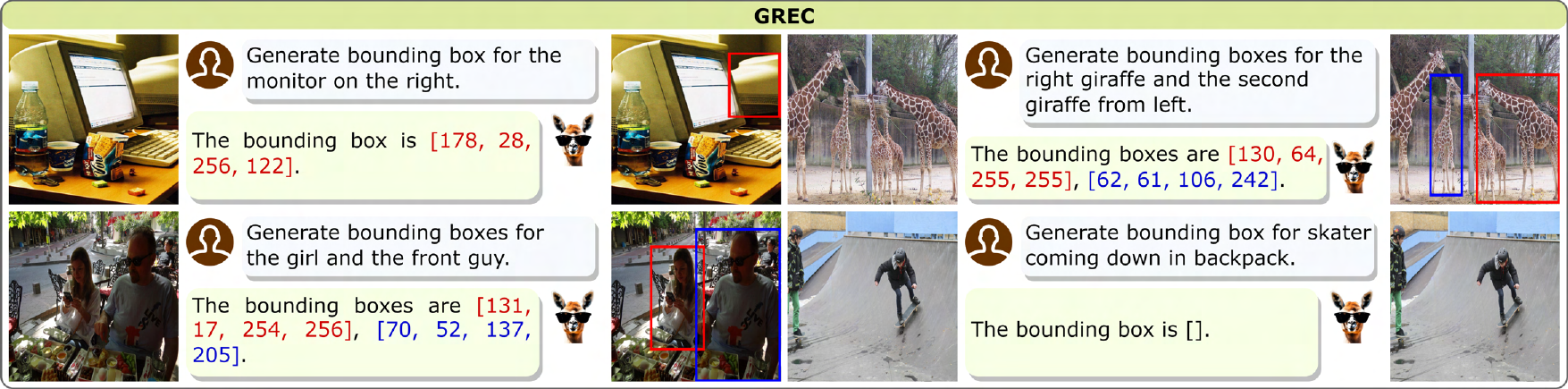}
    \vspace{-6mm}
    \caption{\textbf{Generalized Referring Expression Comprehension (GREC) on gRefCOCO by \model-13B.} GREC aims to identify all objects described by a referring expression and draw bounding boxes around every referred object. GREC also contains no-target expressions where the output is empty.}
    \label{fig:appendix_grec}
\end{figure*}

\begin{figure*}[!t]
    \centering
    \includegraphics[width=\textwidth]{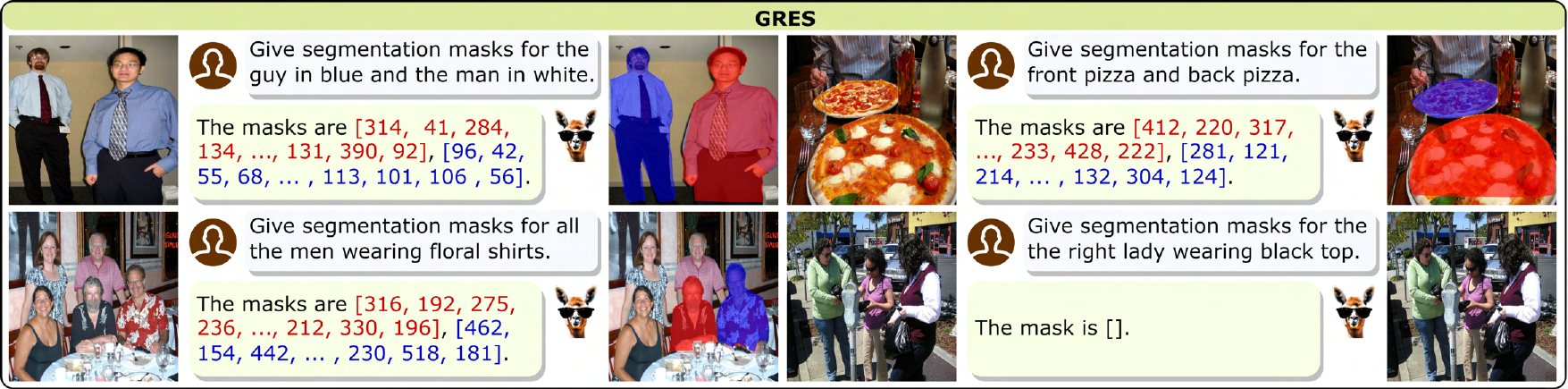}
    \vspace{-6mm}
    \caption{\textbf{Generalized Referring Expression Segmentation (GRES) on gRefCOCO by \model-13B.} GRES aims to identify all objects described by a referring expression and segment every referred object. GRES also contains no-target samples where the output is empty.}
    \label{fig:appendix_gres}
\end{figure*}

\begin{figure*}[!t]
    \centering
    \includegraphics[width=\textwidth]{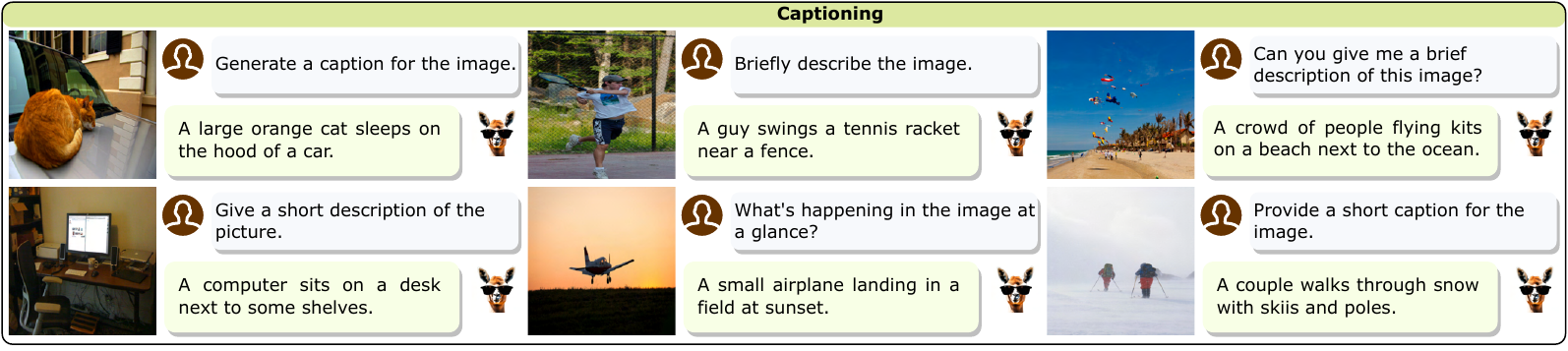}
    \vspace{-6mm}
    \caption{\textbf{Image Captioning on COCO by \model-13B}, which aims to generate a short holistic description of the input image.}
    \label{fig:appendix_caption}
\end{figure*}

\begin{figure*}[!t]
    \centering
    \includegraphics[width=\textwidth]{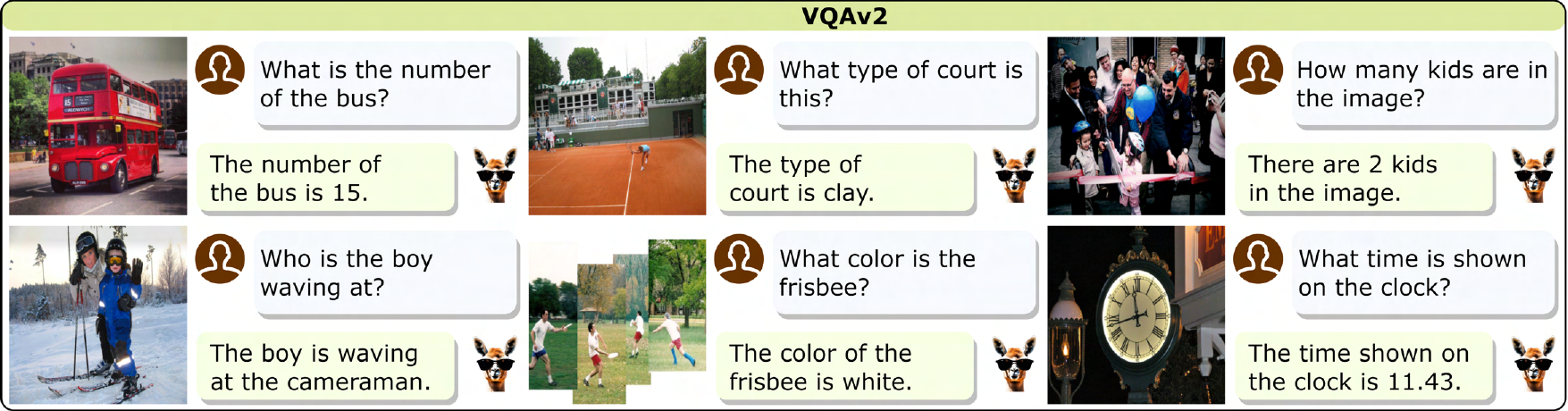}
    \vspace{-6mm}
    \caption{\textbf{VQAv2 by \model-13B}, which aims to answer direct questions based on an input image.}
    \label{fig:appendix_vqav2}
\end{figure*}

\begin{figure*}[!t]
    \centering
    \includegraphics[width=\textwidth]{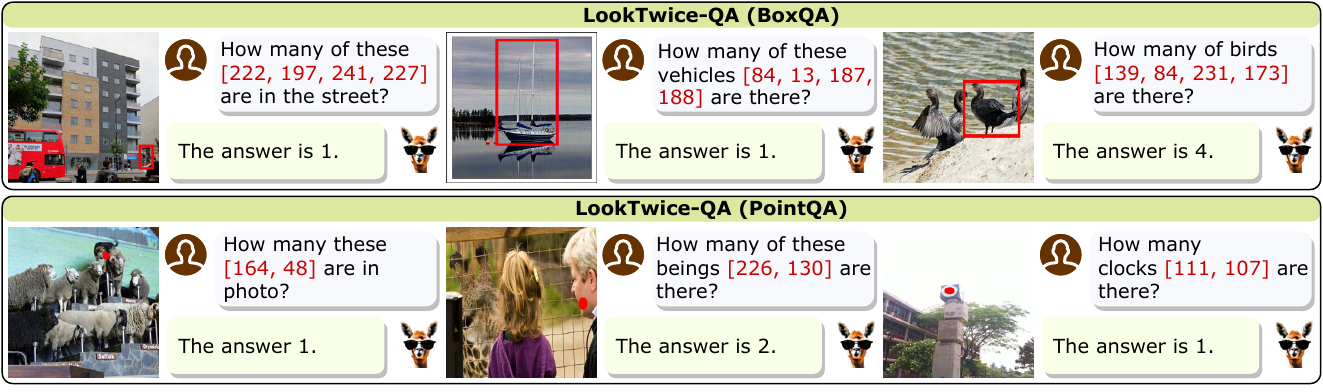}
    \vspace{-6mm}
    \caption{\textbf{Box Question Answering (BoxQA) and Point Question Answering (PointQA) on LookTwice-QA by \model-13B.} Given a question about a specified region in the image, either mentioning a point or a box, this task needs to comprehend the area in the context of the whole image to produce the correct answer.}
    \label{fig:appendix_looktwiceqa}
\end{figure*}

\begin{figure*}[!t]
    \centering
    \includegraphics[width=\textwidth]{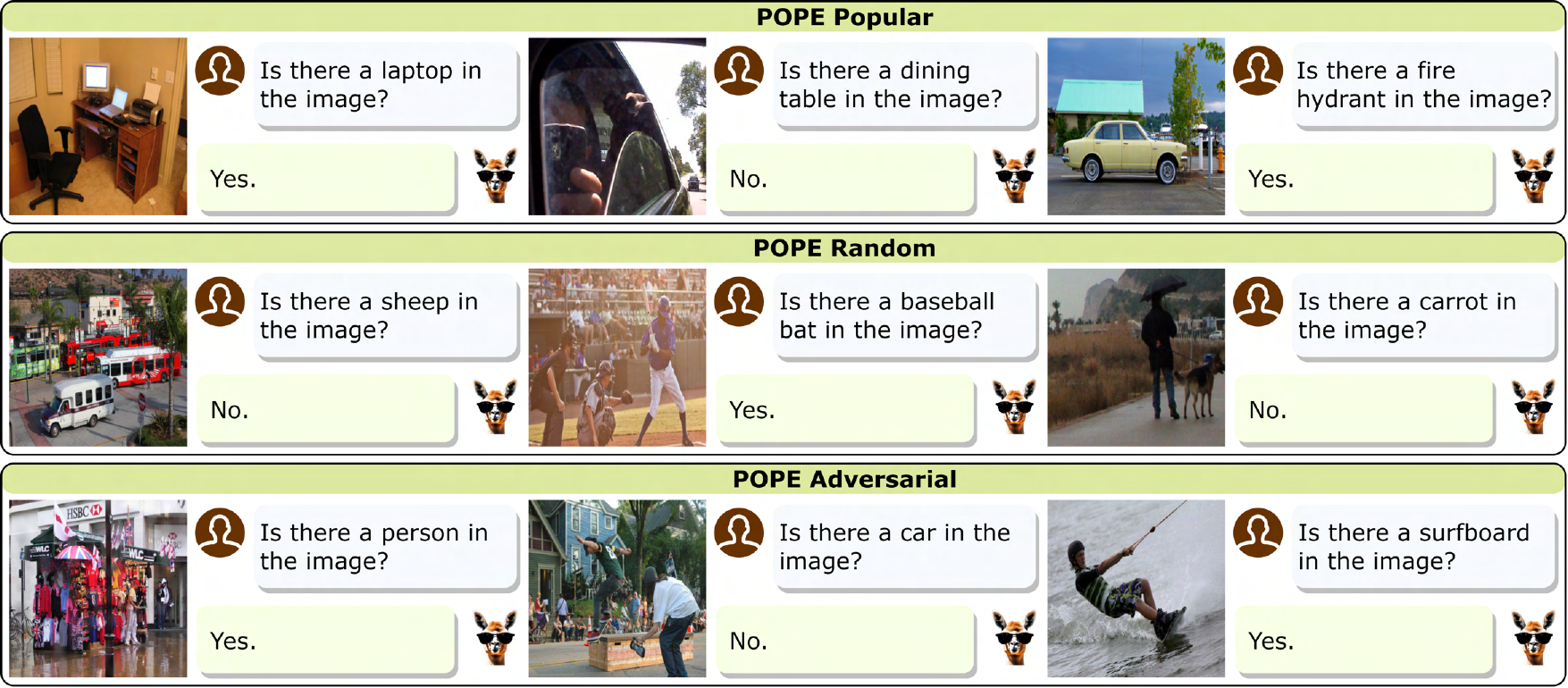}
    \vspace{-6mm}
    \caption{\textbf{Object Hallucination Evaluation of \model-13B on POPE benchmark.} The task aims to input a query inquiring about the existence of an object, and the model is expected to generate a response in the form of either “yes/no.”}
    \label{fig:appendix_pope}
\end{figure*}

\begin{figure*}[!t]
    \centering
    \includegraphics[width=\textwidth]{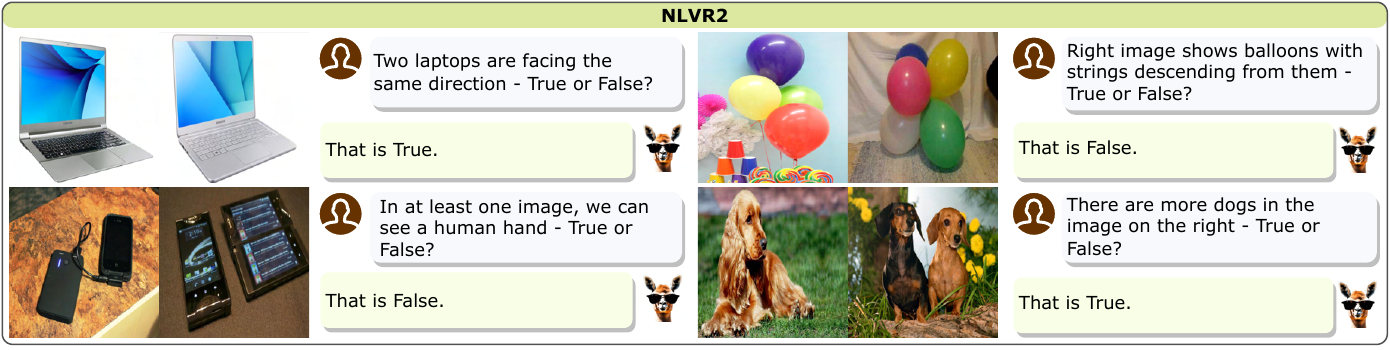}
    \vspace{-6mm}
    \caption{\textbf{Natural Language for Visual Reasoning (NLVR2) by \model-13B.} Given a pair of input images and a question, the model must reason both images to produce the answer correctly.}
    \label{fig:appendix_nlvr2}
\end{figure*}

\begin{figure*}[!t]
    \centering
    \includegraphics[width=\textwidth]{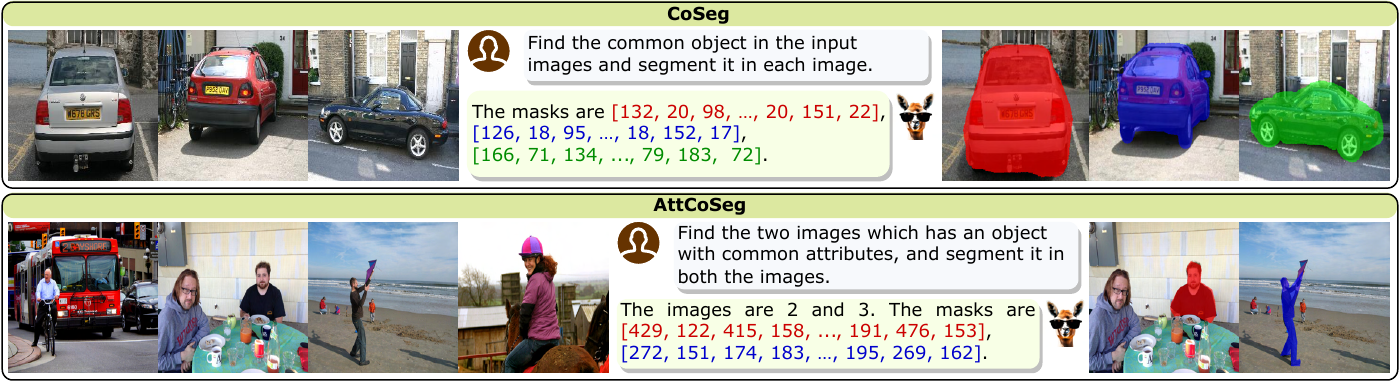}
    \vspace{-6mm}
    \caption{\textbf{CoSeg and AttCoSeg by \model-13B.} Given a set of input images, CoSeg aims to find and segment a common object in every image. AttCoSeg is the more challenging scenario where the input images contains a common object with similar attributes. \model\ is expected to segment the object in both images.}
    \label{fig:appendix_coseg_attcoseg}
\end{figure*}

\begin{figure*}[!t]
    \centering
    \includegraphics[width=\textwidth]{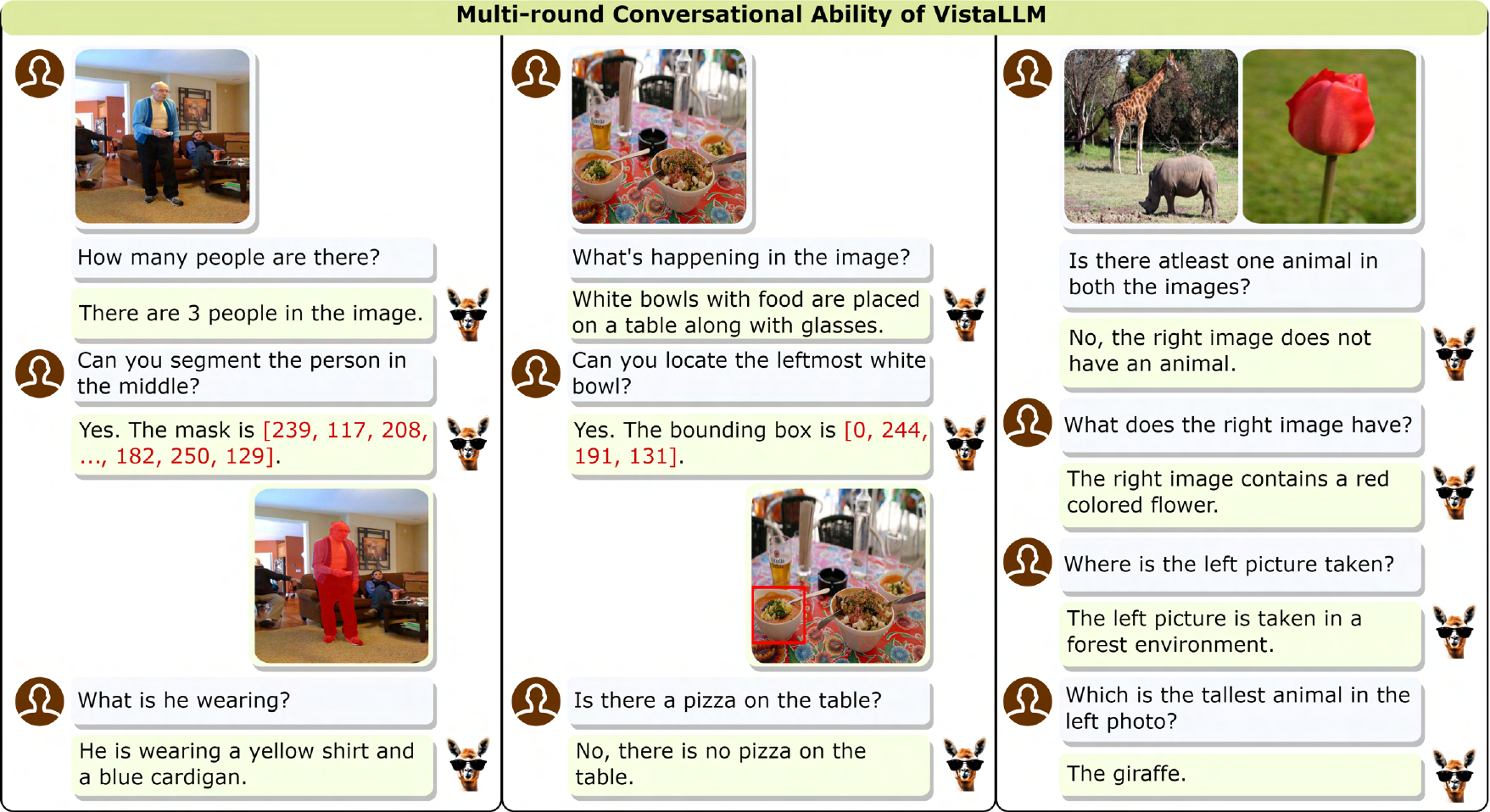}
    \vspace{-6mm}
    \caption{\textbf{Multi-round Conversational Ability of \model-13B.} The images are taken from COCO. \model\ can address all possible grounding and reasoning tasks across single and multiple input images.}
    \label{fig:appendix_conversation}
\end{figure*}

We provide additional qualitative results from \model-13B in Figures \ref{fig:appendix_rec}, \ref{fig:appendix_res}, \ref{fig:appendix_grec}, \ref{fig:appendix_gres}, \ref{fig:appendix_caption}, \ref{fig:appendix_vqav2}, \ref{fig:appendix_looktwiceqa}, \ref{fig:appendix_pope}, \ref{fig:appendix_nlvr2}, and \ref{fig:appendix_coseg_attcoseg}. Moreover, we illustrate multi-round conversational ability of \model\ in Figure \ref{fig:appendix_conversation}.

\end{document}